\documentclass[10pt,twocolumn,letterpaper]{article}

\usepackage[pagenumbers]{cvpr} %

\usepackage{amssymb}
\usepackage{pifont}

\usepackage{booktabs}       %
\usepackage{amsfonts}       %
\usepackage{nicefrac}       %
\usepackage{pifont}
\usepackage{xcolor}         %
\usepackage{multirow}
\usepackage{microtype}      %
\usepackage{graphicx} 
\usepackage{amsmath}
\usepackage{amssymb}
\usepackage{wrapfig}
\usepackage{colortbl}
\usepackage{leftindex}
\usepackage{comment}
\usepackage{url}

\DeclareMathOperator*{\argmin}{arg\,min}

\def\etal{\emph{et al}.}

\newcommand{\myParagraph}[1]{\noindent \textbf{#1}}

\newcommand{\C}[2]{C^{#1,#2}} %
\newcommand{\X}[2]{X^{#1,#2}} %
\newcommand{\Xgt}[2]{\bar{X}^{#1,#2}} %
\newcommand{\z}{z}           %
\newcommand{\zgt}{\bar{z}}   %
\newcommand{\Z}{\text{norm}} %

\newcommand{\F}{\mathcal{F}} %
\newcommand{\D}{\mathcal{D}} %
\newcommand{\lreg}{\ell_{\text{regr}}} %
\newcommand{\lconf}{\mathcal{L}_{\text{conf}}} %

\newcommand{\N}[3]{\text{NN}_{#3}^{#1,#2}} %
\newcommand{\M}{\mathcal{M}} %

\newcommand{\G}{\mathcal{G}} %
\newcommand{\V}{\mathcal{V}} %
\newcommand{\E}{\mathcal{E}} %

\definecolor{Gray}{gray}{0.85}
\definecolor{GrayBorder}{gray}{0.65}
\definecolor{green_cylinder}{rgb}{0.0,0.40,0.0}
\definecolor{blue_cylinder}{rgb}{0.02,0.05,0.75}
\definecolor{way_point}{rgb}{0.56,0.0,1.0}
\newcolumntype{a}{>{\columncolor{GrayBorder}}c}

\newcommand{\duster}{DUSt3R}
\newcommand{\DUSTER}{\textbf{DUSt3R}}
\newcommand{\suppl}{appendix} %
\newcommand{\supplementary}{appendix} %
\newcommand{\supplvideo}{attached video} %

\definecolor{Gray}{gray}{0.90}
\newcolumntype{a}{>{\columncolor{Gray}}c}
\newcolumntype{b}{>{\columncolor{white}}c}

\definecolor{cvprblue}{rgb}{0.21,0.49,0.74}
\usepackage[pagebackref,breaklinks,colorlinks,citecolor=cvprblue]{hyperref}

\def\paperID{4520} %
\def\confName{CVPR}
\def\confYear{2024}

\title{DUSt3R: Geometric 3D Vision Made Easy}

\author{
Shuzhe Wang$^*$, Vincent Leroy$^\dagger$,  Yohann Cabon$^\dagger$,  Boris Chidlovskii$^\dagger$ and Jerome Revaud$^\dagger$\\
\quad$^*$Aalto University\quad\quad\quad\quad\quad\quad\quad\quad$^\dagger$Naver Labs Europe\\
\quad\quad\quad{\tt\small shuzhe.wang@aalto.fi}\quad\quad\quad\quad{\tt\small firstname.lastname@naverlabs.com}
}

\begin{document}

\twocolumn[{%
\renewcommand\twocolumn[1][]{#1}%
\maketitle
\begin{center}
    \captionsetup{type=figure}
    \vspace{-0.7cm}
    \includegraphics[width=0.75\linewidth,trim=0bp 360bp 0bp 0bp,clip]{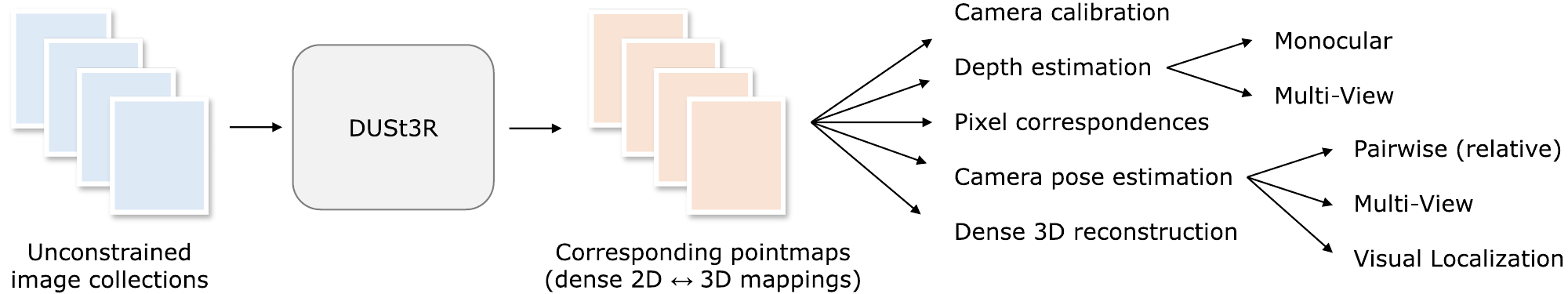} \\
    $\vcenter{\hbox{\resizebox{0.47\linewidth}{!}{\begin{tabular}[c]{@{}c@{}}
    \includegraphics[width=\linewidth]{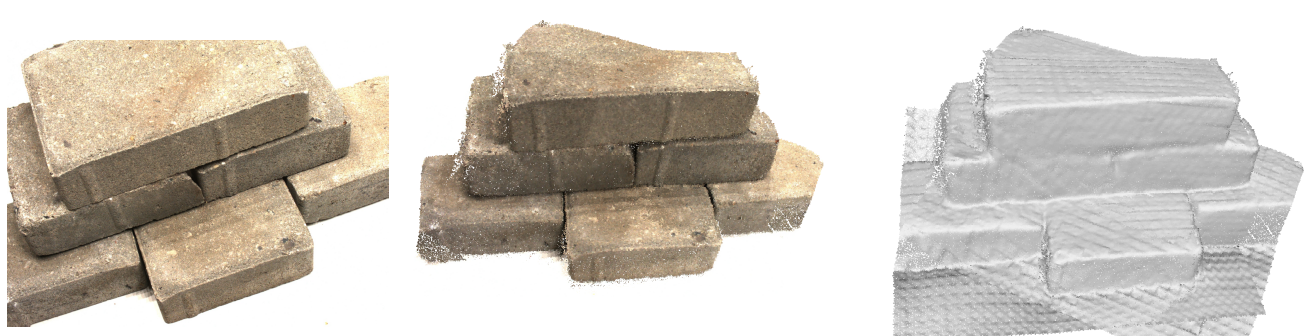} \\
    \includegraphics[width=\linewidth]{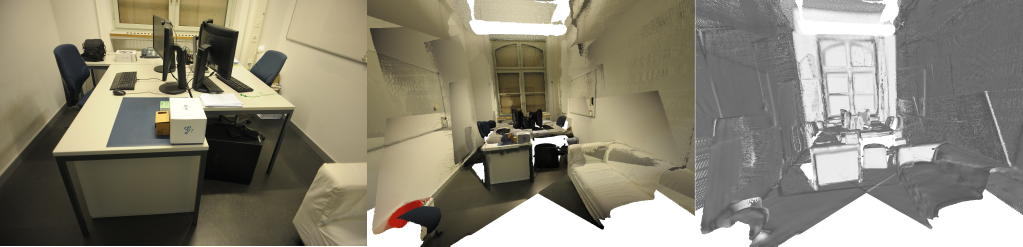} \\
    \includegraphics[width=\linewidth]{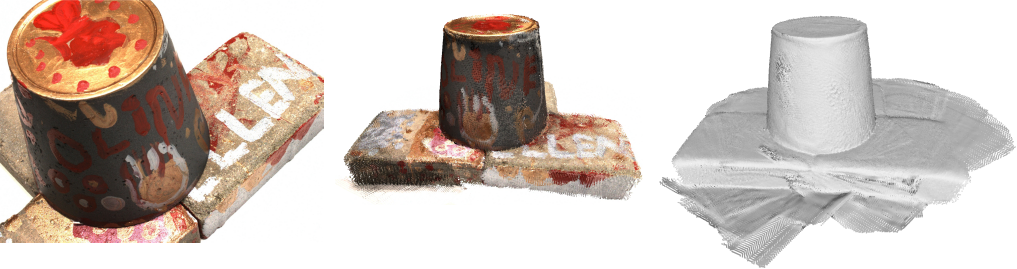} \\
    \end{tabular}}}}$
    $\vcenter{\hbox{\includegraphics[width=0.5\linewidth]{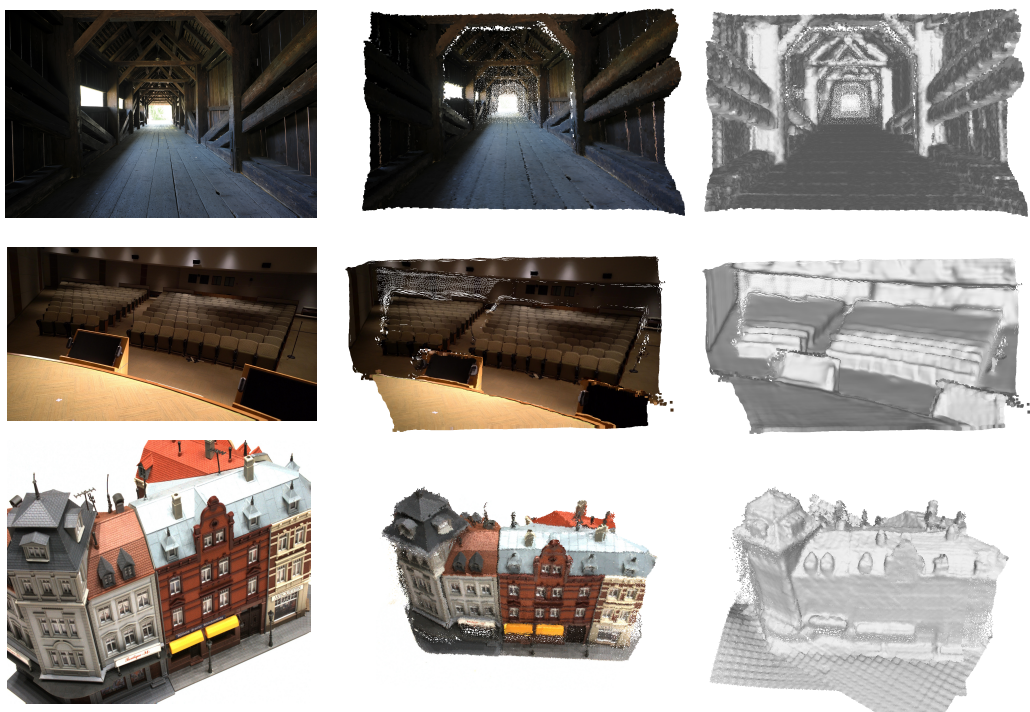}}}$
    \\[-0.4cm]
    \captionof{figure}{
    \textbf{Overview}: 
    Given an unconstrained image collection, \ie a set of photographs with unknown camera poses and intrinsics, our proposed method \DUSTER{} outputs a set of corresponding \emph{pointmaps}, 
    from which we can straightforwardly recover a variety of geometric quantities normally difficult to estimate all at once, such as the camera parameters, pixel correspondences, depthmaps, and fully-consistent 3D reconstruction. 
    Note that \duster{} also works for a single input image (\eg achieving in this case monocular reconstruction).
    We also show \textbf{qualitative examples} on the DTU, Tanks and Temples and ETH-3D datasets~\cite{eth3d_bench,tandt_bench,dtu} obtained \textbf{without} known camera parameters. For each sample, from \emph{left} to \emph{right}: input image, colored point cloud, and rendered with shading for a better view of the underlying geometry.
    }
    \label{fig:qualiCC}
\end{center}
}]

\maketitle
\begin{abstract}
\vspace{-.3cm}
Multi-view stereo reconstruction (MVS) in the wild requires to first estimate the camera parameters e.g. intrinsic and extrinsic parameters. These are usually tedious and cumbersome to obtain, yet they are mandatory to triangulate corresponding pixels in 3D space, which is the core of all best performing MVS algorithms.  
In this work, we take an opposite stance and introduce \textbf{\duster{}}\footnote[1]{\url{https://dust3r.europe.naverlabs.com}}, a radically novel paradigm for \underline{D}ense and \underline{U}nconstrained \underline{S}tereo \underline{3}D \underline{R}econstruction of arbitrary image collections, i.e. operating without prior information about camera calibration nor viewpoint poses. 
We cast the pairwise reconstruction problem as a regression of pointmaps, relaxing the hard constraints of usual projective camera models. We show that this formulation smoothly unifies the monocular and binocular reconstruction cases. 
In the case where more than two images are provided, we further propose a simple yet effective global alignment strategy that expresses all pairwise pointmaps in a common reference frame. 
We base our network architecture on standard Transformer encoders and decoders, allowing us to leverage powerful pretrained models.  
Our formulation directly provides a 3D model of the scene as well as depth information, but interestingly, we can seamlessly recover from it, pixel matches, relative and absolute cameras.   
Exhaustive experiments on all these tasks showcase that the proposed \duster{} can unify various 3D vision tasks and set new SoTAs on monocular/multi-view depth estimation as well as relative pose estimation. 
In summary, \duster{} makes many geometric 3D vision tasks easy.

\end{abstract}
  
\vspace*{-5mm} 
\section{Introduction}
\label{sec:intro}
Unconstrained image-based dense 3D reconstruction from multiple views is one of a few long-researched end-goals of computer vision~\cite{dame2013dense,pointnet,nerf}.
In a nutshell, the task aims at estimating the 3D geometry and camera parameters of a particular scene, given a set of photographs of this scene.
Not only does it have numerous applications like mapping~\cite{mur2015orb, orb-slam3} , navigation~\cite{chaplot2020learning}, archaeology~\cite{peppa2018archaeological, thrun2002probabilistic}, cultural heritage preservation~\cite{culturalheritage}, robotics~\cite{ozyecsil2017survey}, but perhaps more importantly, it holds a fundamentally special place among all 3D vision tasks.
Indeed, it subsumes nearly all of the other geometric 3D vision tasks.
Thus, modern approaches for 3D reconstruction consists in assembling the fruits of decades of advances in various sub-fields such as keypoint detection~\cite{sift,r2d2,superpoint,dusmanu19} and matching~\cite{superglue,sun2021loftr,brachmann19neural,lindenberger2023lightglue}, robust estimation~\cite{barath2023affineglue,brachmann19neural,zhao21progressive}, %
Structure-from-Motion (SfM) and Bundle Adjustment (BA)~\cite{disco2013pami, colmapsfm, pixsfm}, dense Multi-View Stereo (MVS)~\cite{colmapmvs,vis-mvsnet,cvp-mvsnet,pathcmatchnet,casmvsnet}, etc.

In the end, modern SfM and MVS pipelines boil down to solving a series of \emph{minimal problems}: matching points, finding essential matrices, triangulating points, sparsely reconstructing the scene, estimating cameras and finally performing dense reconstruction.
Considering recent advances, this rather complex chain is of course a viable solution in some settings~\cite{unisurf,neus,geoneus,idr,neat,hfneus,nerfingmvs}, yet we argue it is quite unsatisfactory: each sub-problem is not solved perfectly and adds noise to the next step, increasing the complexity and the engineering effort required for the pipeline to work as a whole. 
In this regard, the absence of communication between each sub-problem is quite telling: it would seem more reasonable if they helped each other, \ie dense reconstruction should naturally benefit from the sparse scene that was built to recover camera poses, and vice-versa. 
On top of that, key steps in this pipeline are brittle and prone to break in many cases~\cite{pixsfm}.
For instance, the crucial stage of SfM that serves to estimate all camera parameters, is typically known to fail in many common situations, \eg when the number of scene views is low~\cite{eth3d}, for objects with non-Lambertian surfaces~\cite{chen22nonlambertian}, in case of insufficient camera motion~\cite{orb-slam3}, etc. 
This is concerning, because in the end, ``an MVS algorithm is only as good as the quality of the input images and camera parameters''~\cite{furu_tuto}.

In this paper, we present \DUSTER{}, a radically novel approach for Dense Unconstrained Stereo 3D Reconstruction from un-calibrated and un-posed cameras. 
The main component is a network that can regress a dense and accurate scene representation solely from a \emph{pair} of images, without prior information regarding the scene nor the cameras (not even the intrinsic parameters). 
The resulting scene representation is based on \emph{3D pointmaps} with  rich properties: they simultaneously encapsulate (a) the scene geometry, (b) the relation between pixels and scene points and (c) the relation between the two viewpoints.
From this output alone, practically all scene parameters (\ie cameras and scene geometry) can be straightforwardly extracted. 
This is possible because our network jointly processes the input images and the resulting 3D pointmaps, thus learning to associate 2D structures with 3D shapes, and having the opportunities of solving multiple minimal problems simultaneously, enabling internal `collaboration' between them.

Our model is trained in a fully-supervised manner using a simple regression loss, leveraging large public datasets for which ground-truth annotations are either synthetically generated~\cite{Savva_2019_ICCV,MIFDB16}, reconstructed from SfM softwares~\cite{megadepth,blendedMVS} or captured using dedicated sensors~\cite{co3d,Sun_2020_CVPR,scannet++,arkitscenes}. 
We drift away from the trend of integrating task-specific modules \cite{ec-sfm}, and instead adopt a fully data-driven strategy based on a generic transformer architecture, not enforcing any geometric constraints at inference, but being able to benefit from powerful pretraining schemes.
The network learns strong geometric and shape priors, which are reminiscent of those commonly leveraged in MVS, like shape from texture, shading or contours~\cite{shan2014occluding}.

To fuse predictions from multiple images pairs, we revisit bundle adjustment (BA) for the case of pointmaps, hereby achieving full-scale MVS.
We introduce a global alignment procedure that, contrary to BA, does not involve minimizing reprojection errors.
Instead, we optimize the camera pose and geometry alignment directly in 3D space, which is fast and shows excellent convergence in practice.
Our experiments show that the reconstructions are accurate and consistent between views in real-life scenarios with various unknown sensors. 
We further demonstrate that the same architecture can handle \emph{real-life} monocular and multi-view reconstruction scenarios seamlessly. %
Examples of reconstructions are shown in~\cref{fig:qualiCC} and in the accompanying \href{https://dust3r.europe.naverlabs.com}{video}.

In summary, our contributions are fourfold.
First, we present the first holistic end-to-end 3D reconstruction pipeline from un-calibrated and un-posed images, that unifies monocular and %
binocular 3D
reconstruction.  
Second, we introduce the pointmap representation for MVS applications, that enables the network to predict the 3D shape in a canonical frame, while preserving the implicit relationship between pixels and the scene. This effectively drops many constraints of the usual perspective camera formulation.
Third, we introduce an optimization procedure to globally align  pointmaps in the context of multi-view 3D reconstruction. 
Our procedure can extract effortlessly all usual intermediary outputs of the classical SfM and MVS pipelines. 
In a sense, our approach unifies all 3D vision tasks and considerably simplifies over  the traditional reconstruction pipeline, making \duster{} seem simple and easy in comparison.
Fourth, we demonstrate promising performance on a range of 3D vision tasks
In particular, our all-in-one model achieves state-of-the-art results on monocular and multi-view depth benchmarks, as well as multi-view camera pose estimation.

\section{Related Work}
\label{sec:related}

For the sake of space, we summarize here the most related works in 3D vision, and refer the reader to the \supplementary{} in \cref{sec:suprel} for a more comprehensive review.

\myParagraph{Structure-from-Motion (SfM)}~\cite{hartleymultiviewgeometry, jiang13, disco2013pami,cui2017hsfm, colmapsfm} aims at reconstructing sparse 3D maps while jointly determining camera parameters from a set of images. 
The traditional pipeline starts from pixel correspondences obtained from  keypoint matching~\cite{harris1988combined, sift, bay2006surf, rosten2006machine, barroso2019key} between multiple images to determine geometric relationships,
followed by bundle adjustment to optimize 3D coordinates and camera parameters jointly. 
Recently, the SfM pipeline has undergone substantial enhancements, particularly with the incorporation of learning-based techniques into its subprocesses. 
These improvements encompass advanced feature description~\cite{yi2016lift, superpoint, dusmanu19, r2d2, disk}, more accurate image matching~\cite{superglue, sun2021loftr, tang2022quadtree, chen2022aspanformer, lindenberger2023lightglue, pautrat_suarez_2023_gluestick, Wang_2023_ICCV, barath2023affineglue}, featuremetric %
refinement~\cite{pixsfm}, and neural bundle adjustment~\cite{barf, xiao2023level}. 
Despite these advancements, the sequential structure of the SfM pipeline persists, %
making it vulnerable
to noise and errors in each individual component.

\myParagraph{MultiView Stereo (MVS)}
is the task of densely reconstructing visible surfaces, which is achieved via triangulation between multiple viewpoints.
In the classical formulation of MVS, all camera parameters are supposed to be provided as inputs.
The fully handcrafted~\cite{furu_tuto,openmvs, gipuma,colmapmvs,apdmvs}, the more recent scene optimization based~\cite{dvr,unisurf,neus,geoneus,idr,neat,hfneus,nerfingmvs}, or learning based~\cite{volsweep,mvsnet,geomvsnet,dmvsnet,unimvsnet,cermvs,casmvsnet} approaches all depend on camera parameter estimates obtained via complex calibration procedures, either during the data acquisition~\cite{dtu,eth3d,scannet,scannet++} or using Structure-from-Motion approaches~\cite{colmapsfm, jiang13} for in-the-wild reconstructions. 
Yet, in real-life scenarios, the inaccuracy of %
pre-estimated camera parameters
can be detrimental for these algorithms to work properly. 
In this work, we propose instead to directly predict the geometry of visible surfaces without any explicit knowledge of the camera parameters.

\myParagraph{Direct RGB-to-3D}. %
Recently, some approaches aiming at directly predicting 3D geometry from a single RGB image have been proposed. 
Since the problem is by nature ill-posed without introducing additional assumptions, these methods leverage neural networks that learn strong 3D priors from large datasets %
to solve ambiguities.
These methods can be classified into two groups. %
The first group leverages class-level object priors.
For instance, Pavllo~\etal~\cite{shapefromsingleimg23,PavlloKHL21,PavlloSHML20} propose to learn a model that can fully recover shape, pose, and appearance from a single image, given a large collection of 2D images. %
While this type of approach is powerful, it does not allow to infer shape on objects from unseen categories.
A second group of work, closest to our method, focuses instead on general scenes.
These methods systematically build on or re-use existing monocular depth estimation (MDE) networks~\cite{dpt,leres21,bian22csdepthv2,yin2022accurate}.
Depth maps indeed encode a form of 3D information and, combined with camera intrinsics, can straightforwardly yield pixel-aligned 3D point-clouds.
SynSin~\cite{synsyn}, for example, performs new viewpoint synthesis from a single image by rendering feature-augmented depthmaps %
knowing all camera parameters.
Without camera intrinsics, %
one solution is to infer them by exploiting temporal consistency in video frames, either by enforcing a global alignment~\etal~\cite{xu2023frozenrecon} or by leveraging differentiable rendering with a photometric reconstruction loss~\cite{spencer2023kick,monodepth}.
Another way is to explicitly learn to predict camera intrinsics, which enables to perform metric 3D reconstruction from a single image when combined with MDE~\cite{metric3d,yin2022towards}.
All these methods are, however, intrinsically limited by the quality of depth estimates, which arguably is ill-posed for monocular settings.

In contrast, our network processes two viewpoints simultaneously in order to output depthmaps, or rather, pointmaps.
In theory, at least, this makes triangulation between rays from different viewpoint possible.
Multi-view networks for 3D reconstruction have been proposed in the past.
They are essentially based on the idea of building a differentiable SfM pipeline, replicating the traditional pipeline but training it end-to-end~\cite{demon,deepv2d,deeptam}. 
For that, however, ground-truth camera intrinsics are required as input, and the output is generally a depthmap and a relative camera pose~\cite{demon,deeptam}.
In contrast, our network has a generic architecture and outputs pointmaps, \ie dense 2D field of 3D points, which handle camera poses implicitly and makes the regression problem much better posed.

\myParagraph{Pointmaps.}
Using a collection of pointmaps as shape representation is quite counter-intuitive for MVS, but its usage is widespread for Visual Localization tasks, either in scene-dependent optimization approaches~\cite{dsac,dsac++,dsac*} or scene-agnostic inference methods~\cite{tang21,sanet,sacreg}. Similarly, view-wise modeling is a common theme in monocular 3D reconstruction works~\cite{shin18,pgcd3d,tatarchenko16,mvpc} and in view synthesis works~\cite{synsyn}. The idea being to store the canonical 3D shape in multiple canonical views to work in image space. These approaches usually leverage explicit perspective camera geometry, via rendering of the canonical representation.

\begin{figure*}
    \centering
    \includegraphics[width=1\linewidth, trim=30 300 20 20,clip]{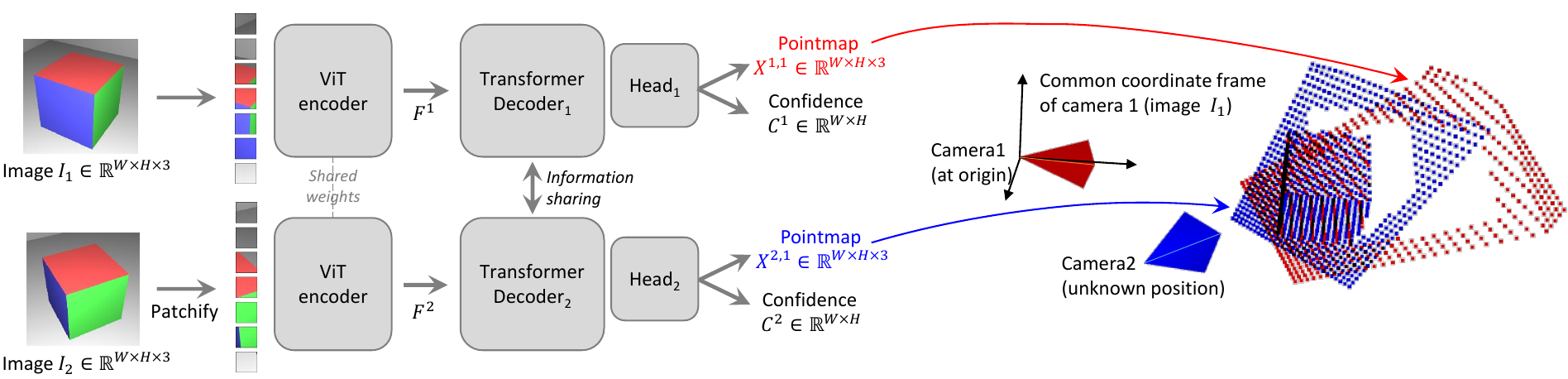}
    \vspace{-6mm}
    \caption{\textbf{Architecture of the network $\F$.} Two views of a scene $(I^1, I^2)$ are first encoded in a Siamese manner with a shared ViT encoder. The resulting token representations $F^1$ and $F^2$ are then passed to two transformer decoders that constantly exchange information via cross-attention.
    Finally, two regression heads output the two corresponding pointmaps and associated confidence maps. 
    Importantly, the two pointmaps are expressed in the same coordinate frame of the first image $I^1$.
    The network $\F$ is trained using a simple regression loss (\cref{eq:conf_loss})
    }
    \label{fig:arch_duster}
    \vspace*{-3mm}
\end{figure*}

\section{Method}
\label{sec:method}

Before delving into the details of our method, we introduce below the essential concept of pointmaps.

\myParagraph{Pointmap}.
In the following, we denote a dense 2D field of 3D points as a \emph{pointmap} $X\in \mathbb{R}^{W\times H \times 3}$.
In association with its corresponding RGB image $I$ of resolution $W \times H$, $X$ forms a one-to-one mapping between image pixels and 3D scene points, \ie $I_{i,j} \leftrightarrow X_{i,j}$, for all pixel coordinates $(i,j) \in \{1\ldots W\}\times\{1\ldots H\}$. %
We assume here that each camera ray hits a single 3D point, \ie ignoring the case of translucent surfaces.

\myParagraph{Cameras and scene}.
Given the camera intrinsics $K\in \mathbb{R}^{3 \times 3}$, the pointmap $X$ of the observed scene can be straightforwardly obtained from the ground-truth depthmap $D\in \mathbb{R}^{W \times H}$ as $X_{i,j} = K^{-1} \left[i D_{i,j}, j D_{i,j}, D_{i,j}\right]^\top$.
Here, $X$ is expressed in the camera coordinate frame. %
In the following, we denote as $\X{n}{m}$ the pointmap $X^n$ from camera $n$ expressed in camera $m$'s coordinate frame:
\vspace{-2mm}
\begin{equation}
    \X{n}{m} = P_m P_n^{-1} h\left( X^n \right)
    \label{eq:pointmap}
\vspace{-2mm}
\end{equation}
with $P_m, P_n\in \mathbb{R}^{3 \times 4}$ 
the world-to-camera poses for images $n$ and $m$, 
and $h:(x,y,z)\rightarrow(x,y,z,1)$ the homogeneous mapping.

\subsection{Overview}
We wish to build a network that solves the 3D reconstruction task for the generalized stereo case through direct regression.
To that aim, we train a network $\F$ that takes as input 2 RGB images $I^1, I^2 \in \mathbb{R}^{W\times H \times 3}$ and outputs 2 corresponding pointmaps $\X{1}{1},\X{2}{1} \in \mathbb{R}^{W\times H \times 3}$ with associated confidence maps $\C{1}{1},\C{2}{1} \in \mathbb{R}^{W\times H}$. 
Note that both pointmaps are %
expressed in the \emph{same} coordinate frame of $I^1$, which radically differs from existing approaches but offers key advantages (see~\cref{sec:intro,sec:related,ssec:postproc,ssec:mvoptim}). 
For the sake of clarity and without loss of generalization, we assume that both images have the same resolution $W\times H$, but naturally in practice their resolution can differ. 

\myParagraph{Network architecture.}
The architecture of our network $\F$ is inspired by CroCo~\cite{croco}, making it straightforward to heavily benefit from CroCo pretraining~\cite{croco_v2}. 
As shown in \cref{fig:arch_duster}, it is composed of two identical branches (one for each image) comprising each an image encoder, a decoder %
and a regression head.
The two input images are first encoded in a Siamese manner by the same weight-sharing ViT encoder~\cite{vit}, yielding two token representations $F^1$ and $F^2$:
\vspace{-2mm}
\[
  F^1 = \text{Encoder}(I^1), F^2 = \text{Encoder}(I^2).
\vspace{-2mm}
\]
The network then reasons over both of them jointly in the decoder. Similarly to CroCo~\cite{croco}, the decoder is a generic transformer network equipped with cross attention. 
Each decoder block thus sequentially performs self-attention (each token of a view attends to tokens of the same view), then cross-attention (each token of a view attends to all other tokens of the other view), and finally feeds tokens to a MLP.
Importantly, information is constantly shared between the two branches during the decoder pass.
This is crucial in order to output properly aligned pointmaps.
Namely, each decoder block attends to tokens from the other branch:
\vspace{-2mm}
\begin{eqnarray*}
  G_i^1 = \text{DecoderBlock}^1_{i}\left(G_{i-1}^1, G_{i-1}^2\right),\\
  G_i^2 = \text{DecoderBlock}^2_{i}\left(G_{i-1}^2, G_{i-1}^1\right),
  \vspace{-3mm}
\end{eqnarray*}
for $i = 1,\ldots,B$ for a decoder with $B$ blocks and initialized with encoder tokens $G_0^1:=F^1$ and $G_0^2:=F^2$. 
Here, $\text{DecoderBlock}_i^v(G^1,G^2)$ denotes the $i$-th block in branch $v\in\{1,2\}$, $G^1$ and $G^2$ are the input tokens, with $G^2$ the tokens from the other branch. 
Finally, in each branch a separate regression head takes the set of decoder tokens and outputs a pointmap and an associated confidence map:
\vspace{-2mm}
\begin{eqnarray*}
  \X{1}{1}, \C{1}{1} = \text{Head}^1\left(G_0^1, \ldots, G_B^1\right), \\
  \X{2}{1}, \C{2}{1} = \text{Head}^2\left(G_0^2, \ldots, G_B^2\right).
\end{eqnarray*}

\myParagraph{Discussion}.
The output pointmaps $\X{1}{1}$ and $\X{2}{1}$ are regressed up to an unknown scale factor.
Also, it should be noted that our generic architecture never explicitly enforces any geometrical constraints.
Hence, pointmaps do not necessarily correspond to any physically plausible camera model.
Rather, we let the network learn all relevant priors present from the train set, which only contains geometrically consistent pointmaps. 
Using a generic architecture allows to leverage strong pretraining technique, ultimately surpassing what existing task-specific architectures can achieve.
We detail the learning process in the next section.

\subsection{Training Objective}
\label{ssec:losses}

\myParagraph{3D Regression loss.}
Our sole training objective is based on regression in the 3D space.
Let us denote the ground-truth pointmaps as $\Xgt{1}{1}$ and $\Xgt{2}{1}$, obtained from \cref{eq:pointmap} along with two corresponding sets of valid pixels $\D^1,\D^2 \subseteq \{1\ldots W\}\times\{1\ldots H\}$ %
on which the ground-truth is defined.
The regression loss for a valid pixel $i\in\D^v$ in view $v\in\{1,2\}$ is simply defined as the Euclidean distance:
\vspace{-2mm}
\begin{equation}
\lreg(v,i) = \left\Vert \frac{1}{\z}\X{v}{1}_{i}  - \frac{1}{\zgt}\Xgt{v}{1}_{i} \right\Vert.
\vspace{-1mm}
\label{eq:regression}
\end{equation}
To handle the scale ambiguity between prediction and ground-truth, we normalize the predicted and ground-truth pointmaps by scaling factors $\z=\Z(\X{1}{1},\X{2}{1})$ and $\zgt=\Z(\Xgt{1}{1},\Xgt{2}{1})$, respectively, which simply represent the average distance of all valid points to the origin:
\vspace{-2mm}
\begin{equation}
\Z(X^1,X^2) = \frac{1}{|\D^1|+|\D^2|} \sum_{v \in \{1,2\}} \, 
              \sum_{i \in \D^v} \left\Vert X^v_{i} \right\Vert.
\vspace{-2mm}
\label{eq:normalization}
\end{equation}

\begin{figure*}
    \centering
    \includegraphics[width=0.56\linewidth,trim=0 270 40 0,clip]{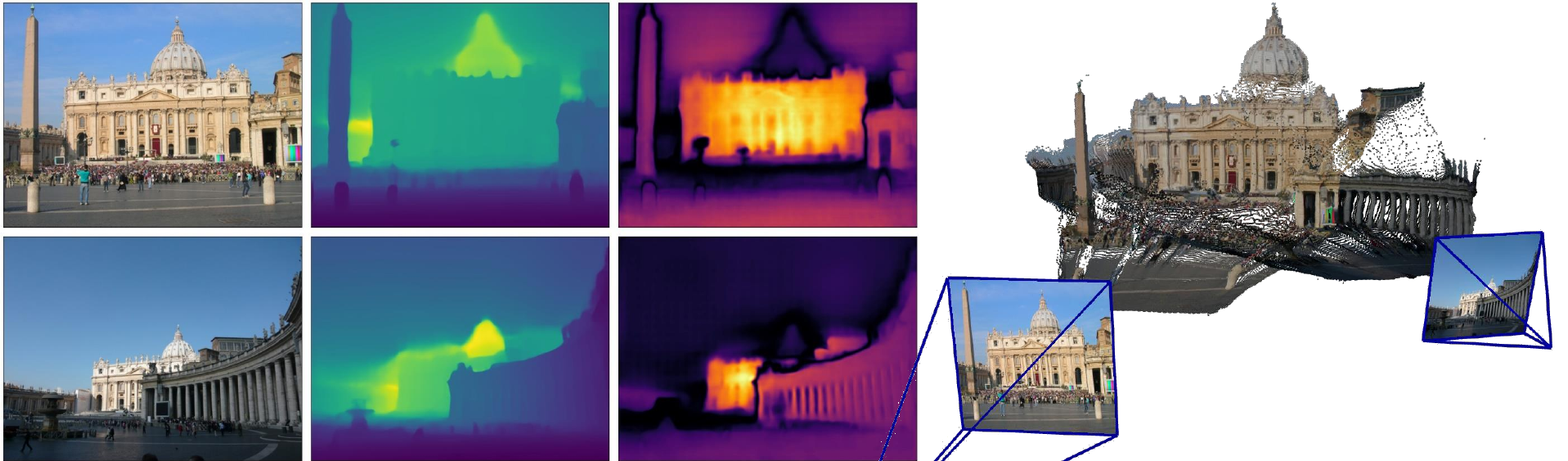}
    \includegraphics[width=0.42\linewidth]{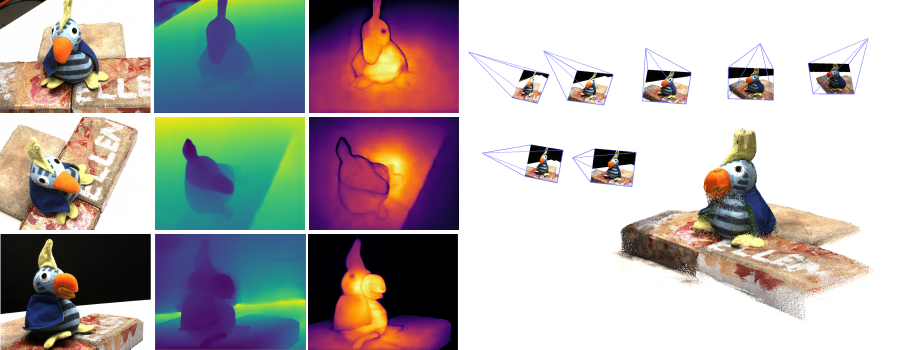}
    \\[-0.2cm]
    \caption{
    \textbf{Reconstruction examples} on two scenes never seen during training. 
    From left to right: RGB, depth map, confidence map, reconstruction. 
    The left scene shows the raw result output from $\F(I^1,I^2)$.
    The right scene shows the outcome of global alignment (\cref{ssec:mvoptim}).
    }
    \label{fig:megadepth_confidence}
    \vspace*{-4mm}  
\end{figure*}

\myParagraph{Confidence-aware loss}.
In reality, and contrary to our assumption, there are ill-defined 3D points, \eg in the sky or on translucent objects.
More generally, some parts in the image are typically harder to predict that others.
We thus jointly learn to predict a score for each pixel which represents the confidence that the network has about this particular pixel.
The final training objective is the confidence-weighted regression loss from \cref{eq:regression} over all valid pixels:
\vspace{-2mm}
\begin{equation}
  \lconf = \sum_{v \in \{1,2\}} \, \sum_{i \in \D^v} 
            \C{v}{1}_i \lreg(v,i) - \alpha \log \C{v}{1}_i,
\label{eq:conf_loss}
\vspace{-2mm}
\end{equation}
where $\C{v}{1}_i$ is the confidence score for pixel $i$, and $\alpha$ is a hyper-parameter controlling the regularization term~\cite{confnet18}. 
To ensure a strictly positive confidence, we typically define 
$\C{v}{1}_i=1+\exp \widetilde{\C{v}{1}_i} > 1$.
This has the effect of forcing the network to extrapolate in harder areas, \eg like those ones covered by a single view.
Training network $\F$ with this objective allows to estimate confidence scores without an explicit supervision.
Examples of input image pairs with their corresponding outputs are shown in~\cref{fig:megadepth_confidence} and in the \suppl{} in~\cref{fig:megadepth_1,fig:megadepth_2,fig:opposite}.

\subsection{Downstream Applications}
\label{ssec:postproc}

The rich properties of the output pointmaps allows us to perform various convenient operations with relative ease.

\myParagraph{Point matching.}
Establishing correspondences between pixels of two images can be trivially achieved by nearest neighbor (NN) search in the 3D pointmap space.
To minimize errors, we typically retain reciprocal (mutual) correspondences $\M_{1,2}$ between images $I^1$ and $I^2$, \ie we have:
\begin{eqnarray*}
  \M_{1,2} = \{(i,j) \ |\  i=\N{1}{2}{1}(j) \text{ and } j=\N{2}{1}{1}(i) \} \\
  \text{with } \N{n}{m}{k}(i) = \argmin_{j \in \{0,\ldots,WH\}} \left\Vert \X{n}{k}_j - \X{m}{k}_i \right\Vert. %
\end{eqnarray*}

\myParagraph{Recovering intrinsics.}
By definition, the pointmap $\X{1}{1}$ is expressed in $I^1$'s coordinate frame.
It is therefore possible to estimate the camera intrinsic parameters by solving a simple optimization problem.
In this work, we assume that the principal point is approximately centered and pixel are squares, hence only the focal $f_1^*$ remains to be estimated:
\vspace{-1mm}
\[
  f_1^* = \argmin_{f_1} \sum_{i=0}^{W} \sum_{j=0}^{H} \C{1}{1}_{i,j} \left\Vert (i',j') - f_1 \frac{(\X{1}{1}_{i,j,0},\X{1}{1}_{i,j,1})}{\X{1}{1}_{i,j,2}} \right\Vert,
  \vspace{-1mm}
\]
with $i'=i-\frac{W}{2}$ and $j'=j-\frac{H}{2}$. 
Fast iterative solvers, \eg based on the Weiszfeld algorithm~\cite{weiszfeld}, can find the optimal $f_1^*$ in a few iterations.
For the focal $f_2^*$ of the second camera, the simplest option is to perform the inference for the pair $(I^2,I^1)$ and use above formula with $\X{2}{2}$ instead of $\X{1}{1}$.

\myParagraph{Relative pose estimation}
can be achieved in several fashions.
One way is to perform 2D matching and recover intrinsics as described above, then estimate the Epipolar matrix and recover the relative pose~\cite{hartleymultiviewgeometry}.
Another, more direct, way is to compare the pointmaps $\X{1}{1} \leftrightarrow \X{1}{2}$ (or, equivalently, $\X{2}{2} \leftrightarrow \X{1}{2}$) using Procrustes alignment~\cite{procrustes} to get the relative pose
$P^{*}=[R^{*}|t^{*}]$:
\vspace{-2mm}
\[
   R^*, t^* = \argmin_{\sigma,R,t} \sum_{i} \C{1}{1}_i \C{1}{2}_i \left\Vert \sigma (R \X{1}{1}_i + t) - \X{1}{2}_i \right\Vert^2,
   \vspace{-2mm}
\]
which can be achieved in closed-form. 
Procrustes alignment is, unfortunately, sensitive to noise and outliers.
A more robust solution is finally to rely on RANSAC~\cite{fischlerrandom1981} with PnP~\cite{hartleymultiviewgeometry,epnp}.

\myParagraph{Absolute pose estimation}, also termed visual localization, can likewise be achieved in several different ways.
Let $I^Q$ denote the query image and $I^B$ the reference image for which 2D-3D correspondences are available.
First, intrinsics for $I^Q$ can be estimated from $\X{Q}{Q}$.
One possibility consists of obtaining 2D correspondences between $I^Q$ and $I^B$, which in turn yields 2D-3D correspondences for $I^Q$, and then running PnP-RANSAC~\cite{fischlerrandom1981,epnp}.
Another solution is to get the relative pose between $I^Q$ and $I^B$ as described previously.
Then, we convert this pose to world coordinate by scaling it appropriately, according to the scale between $\X{B}{B}$ and the ground-truth pointmap for $I^B$. %

\subsection{Global Alignment}
\label{ssec:mvoptim}

The network $\F$ presented so far can only handle a pair of images.
We now present a fast and simple post-processing optimization for entire scenes that enables the alignment of pointmaps predicted from multiple images into a joint 3D space.
This is possible thanks to the rich content of our pointmaps, which encompasses by design two aligned point-clouds and their corresponding pixel-to-3D mapping.

\myParagraph{Pairwise graph.}
Given a set of images $\{I^1, I^2,\ldots, I^N\}$ for a given scene, we first construct a connectivity graph $\G(\V,\E)$ where $N$ images form vertices $\V$ and each edge $e=(n,m) \in \E$ indicates that images $I^n$ and $I^m$ shares some visual content.
To that aim, we either use existing off-the-shelf image retrieval methods, or we pass all pairs through network $\F$ (inference takes $\approx$40ms on a H100 GPU) and measure their overlap based on the average confidence in both pairs, then we filter out low-confidence pairs.

\myParagraph{Global optimization.}
We use the connectivity graph $\G$ to recover \emph{globally aligned} pointmaps $\{\chi^n\in \mathbb{R}^{W\times H\times 3}\}$ for all cameras $n=1\ldots N$.
To that aim, we first predict, for each image pair $e=(n,m) \in \E$, the pairwise pointmaps $\X{n}{n}, \X{m}{n}$ and their associated confidence maps $\C{n}{n},\C{m}{n}$.
For the sake of clarity, let us define $\X{n}{e}:=\X{n}{n}$ and $\X{m}{e}:=\X{m}{n}$.
Since our goal involves to rotate all pairwise predictions in a common coordinate frame, we introduce a pairwise pose $P_e\in\mathbb{R}^{3\times4}$ and scaling $\sigma_e>0$ associated to each pair $e \in \E$.
We then formulate the following optimization problem:
\vspace{-3mm}
\begin{eqnarray}
  \chi^* = \argmin_{\chi,P,\sigma} \sum_{e \in \E} \sum_{v \in e} \sum_{i=1}^{HW}
    \C{v}{e}_i \left\Vert \chi_i^v - \sigma_e P_e \X{v}{e}_i \right\Vert.
\label{eq:pose_optim}
\vspace{-3mm}
\end{eqnarray}
Here, we abuse notation and write $v\in e$ for $v \in \{n,m\}$ if $e=(n,m)$.
The idea is that, for a given pair $e$, the \emph{same} rigid transformation $P_e$ should align both pointmaps $\X{n}{e}$ and $\X{m}{e}$ with the world-coordinate pointmaps $\chi^n$ and $\chi^m$, 
since $\X{n}{e}$ and $\X{m}{e}$ are by definition both expressed in the same coordinate frame.
To avoid the trivial optimum where $\sigma_e=0,\, \forall e\in\E$, we enforce that $\prod_e \sigma_e=1$.

\myParagraph{Recovering camera parameters.}
A straightforward extension to this framework enables to recover all cameras parameters.
By simply replacing $\chi^n_{i,j}:=P_n^{-1}h(K_n^{-1}[iD^n_{i,j};jD^n_{i,j};D^n_{i,j}])$ (\ie enforcing a standard camera pinhole model as in~\cref{eq:pointmap}), we can thus estimate all camera poses $\{P_n\}$,  associated intrinsics $\{K_n\}$ and depthmaps $\{D^n\}$ for $n=1\ldots N$.

\myParagraph{Discussion.}
We point out that, contrary to traditional bundle adjustment, this global optimization is fast and simple to perform in practice. %
Indeed, we are not minimizing 2D reprojection errors, as bundle adjustment normally does, but 3D projection errors.
The optimization is carried out using standard gradient descent and typically converges after a few hundred steps, requiring mere seconds on a standard GPU.

\section{Experiments with \duster}
\label{sec:dustereval}

\myParagraph{Training data}.
We train our network with a mixture of eight datasets:
Habitat~\cite{Savva_2019_ICCV}, MegaDepth~\cite{megadepth}, ARKitScenes~\cite{arkitscenes}, MegaDepth~\cite{megadepth}, Static Scenes 3D~\cite{robust_mvd}, Blended MVS~\cite{blendedMVS}, ScanNet++~\cite{scannet++}, CO3D-v2~\cite{co3d} and Waymo~\cite{Sun_2020_CVPR}. 
These datasets feature diverse scenes types: indoor, outdoor, synthetic, real-world, object-centric, etc. 
When image pairs are not directly provided with the dataset, we extract them based on the method described in~\cite{croco_v2}. Specifically, we utilize off-the-shelf image retrieval and point matching algorithms to match and verify image pairs. 
All in all, we extract 8.5M pairs in total.

\myParagraph{Training details}.
During each epoch, we randomly sample an equal number of pairs from each dataset to equalize disparities in dataset sizes. 
We wish to feed relatively high-resolution images to our network, say 512 pixels in the largest dimension. 
To mitigate the high cost associated with such input, we train our network sequentially, first on 224$\times$224 images and then 
on larger 512-pixel images.
We randomly select the image aspect ratios for each batch (\eg 16/9, 4/3, etc), so that at test time our network is familiar with different image shapes.
We simply crop images to the desired aspect-ratio, and resize so that the largest dimension is 512 pixels. 

We use standard data augmentation techniques and training set-up overall.
Our network architecture comprises a Vit-Large for the encoder~\cite{vit}, a ViT-Base for the decoder and a DPT head~\cite{dpt}.
We refer to the \supplementary{} in \cref{sec:training_details} for more details on the training and architecture.
Before training, we initialize our network with the weights of an off-the-shelf CroCo pretrained model~\cite{croco_v2}.
Cross-View completion (CroCo) is a recently proposed pretraining paradigm inspired by MAE~\cite{mae} that has been shown to excel on various downstream 3D vision tasks, and is thus particularly suited to our framework.
We ablate in \cref{ssec:ablations} the impact of CroCo pretraining and increase in image resolution.

\myParagraph{Evaluation}.
In the remainder of this section, we benchmark \duster{} on a representative set of classical 3D vision tasks, each time specifying datasets, metrics and comparing performance with existing state-of-the-art approaches.
We emphasize that all results are obtained with the \emph{same} \duster{} model (our default model is denoted as `\duster{} 512', other \duster{} models serves for the ablations in Section~\cref{ssec:ablations}), \ie we never finetune our model on a particular downstream task.
During test, all test images are rescaled to 512px while preserving their aspect ratio.
Since there may exist different `routes' to extract task-specific outputs from \duster{}, as described in \cref{ssec:postproc} and \cref{ssec:mvoptim}, we precise each time the employed method.

\myParagraph{Qualitative results}.
\duster{} yields high-quality dense 3D reconstructions even in challenging situations.
We refer the reader to the \supplementary{} in \cref{sec:quali} for non-cherrypicked visualizations of pairwise and multi-view reconstructions.

\vspace{-2mm}
\subsection{Visual Localization}
\begin{table*}[t!] 
\begin{center}
\small
\renewcommand\arraystretch{1.2}
\resizebox{0.8\textwidth}{!}{
\setlength{\tabcolsep}{2pt}

\begin{tabular}{llccccccccccccc}
\hline
\specialrule{1.5pt}{0.5pt}{0.5pt} 
\multicolumn{2}{l}{\multirow{2}{*}{Methods}} & \multicolumn{7}{c}{7Scenes (Indoor)~\cite{7scenes}}                                              &  & \multicolumn{5}{c}{Cambridge (Outdoor)~\cite{cambridge}}                       \\ \cline{3-9} \cline{11-15} 
\multicolumn{2}{l}{}                         & Chess     & Fire      & Heads     & Office    & Pumpkin   & Kitchen   & Stairs    &  & S. Facade & O. Hospital & K. College & St.Mary’s & G. Court   \\ \specialrule{1.5pt}{0.5pt}{0.5pt} 
\multirow{2}{*}{\rotatebox{90}{FM}}      & AS~\cite{SattlerPriorityLoc2017}     & 4/1.96 & 3/1.53 & 2/1.45 & 9/3.61 & 8/3.10 & 7/3.37 & \textbf{3}/2.22 &  & \textbf{4}/0.21    & 20/0.36     & 13/0.22    & 8/0.25    & 24/\bf{0.13}    \\
                         & HLoc ~\cite{hfnet}                & \bf{2/0.79} & \bf{2/0.87} & \bf{2/0.92} & \bf{3/0.91} & \bf{5/1.12} & \bf{4/1.25} & 6/\bf{1.62} &  & \bf{4/0.2}  & \bf{15/0.3}    & \bf{12/0.20}  & \bf{7/0.21} & \textbf{11}/0.16  \\ \hline
\multirow{7}{*}{\rotatebox{90}{E2E}}     & DSAC*~\cite{dsac*}             & \textbf{2}/1.10 & \textbf{2}/1.24 & \textbf{1}/1.82 & \textbf{3}/1.15 & \textbf{4}/1.34 & 4/1.68 & \textbf{3}/1.16 &  & \textbf{5}/0.3  & \textbf{15/0.3}    & 15/0.3   & 13/0.4  & 49/0.3   \\
                             & HSCNet~\cite{li2020hscnet}           & \textbf{2/0.7}  & \textbf{2}/0.9  & \textbf{1}/0.9  & \textbf{3/0.8}  & \textbf{4/1.0}  & 4/\textbf{1.2}  & \textbf{3}/\textbf{0.8}  &  & 6/0.3  & 19/\textbf{0.3}    & 18/0.3   & 9/0.3  & 28/0.2   \\
                         & PixLoc~\cite{sarlin21pixloc}            & \textbf{2}/0/80    & \textbf{2/0.73}    & \textbf{1/0.82}    & \textbf{3}/0.82    & \textbf{4}/1.21    & \textbf{3/1.20}    & 5/1.30    &  & \bf{5/0.23} & 16/0.32   & 14/0.24  & 10/0.34 & 30/0.14  \\
                         & SC-wLS~\cite{sc_wls_eccv2022}           & 3/0.76 & 5/1.09 & 3/1.92 & 6/0.86 & 8/1.27 & 9/1.43 & 12/2.80 &  & 11/0.7  & 42/1.7    & 14/0.6   & 39/1.3  & 164/0.9   \\
                         & NeuMaps ~\cite{neumap}          & \textbf{2}/0.81 & 3/1.11 & 2/1.17 & \textbf{3}/0.98 & \textbf{4}/1.11 & 4/1.33 & 4/1.12 &  & 6/0.25 & 19/0.36   & 14/0.19  & 17/0.53 & \textbf{6/ 0.10} \\ \cline{2-15} 
                         &{\bf \duster{} 224-NoCroCo }               & 5/1.76 & 6/2.02 & 3/1.75 & 5/1.54 & 9/2.35 & 6/1.82 & 34/7.81 &  & 24/1.33 & 79/1.17   & 69/1.15  & 46/1.51 & 143/1.32  \\
                         &{\bf \duster \ 224}               & 3/0.96 & 3/1.02 & \textbf{1}/1.00 & 4/1.04 & 5/1.26 & 4/1.36 & 21/4.08 &  & 9/0.38 & 26/0.46   & 20/0.32  & 11/0.38 & 36/0.24  \\
                         &{\bf \duster \ 512}              & 3/0.97 & 3/0.95 & 2/1.37 & \textbf{3}/1.01 & \textbf{4}/1.14 & 4/1.34 & 11/2.84 &  & 6/0.26 & 17/0.33   & \textbf{11/0.20}& \textbf{7/0.24} & 38/0.16  \\ \specialrule{1.5pt}{0.5pt}{0.5pt} 
\end{tabular}
}

\caption{Absolute camera pose on 7Scenes~\cite{7scenes} and Cambridge-Landmarks~\cite{cambridge} datasets. We report the median translation and rotation errors ($cm/ ^{\circ}$) to feature matching (FM) based and end-to-end (E2E) learning-base methods. The best results at each category are in \bf{bold}.}
\label{tab:cambridge}
\normalsize
\vspace*{-8mm}
\end{center}
\end{table*}

\label{ssec:duster_vl}

\myParagraph{Datasets and metrics.}
We first evaluate \duster{} for the task of absolute pose estimation on the 7Scenes~\cite{7scenes} and Cambridge Landmarks datasets~\cite{cambridge}.
7Scenes contains 7 indoor scenes with RGB-D images from videos and their 6-DOF camera poses.
Cambridge-Landmarks contains 6 outdoor scenes with RGB images and their associated camera poses, which are obtained via SfM.
We report the median translation and rotation errors in (cm/$^{\circ}$), respectively.

\myParagraph{Protocol and results.}
To compute camera poses in world coordinates, we use \duster{} as a 2D-2D pixel matcher (see Section~\ref{ssec:postproc}) between a query and the most relevant database images obtained using off-the-shelf image retrieval AP-GeM~\cite{apgem}. 
In other words, we simply use the raw pointmaps output from $\F(I^Q,I^B)$ without any refinement, where $I^Q$ is the query image and $I^B$ is a database image.
We use the top 20 retrieved images for Cambridge-Landmarks and top 1 for 7Scenes and leverage the known query intrinsics. 
For results obtained without using ground-truth intrinsics parameters, refer to the \supplementary{} in \cref{sec:visloc}.

We compare our results against the state of the art in Table~\ref{tab:cambridge} for each scene of the two datasets. 
Our method obtains comparable accuracy compared to existing approaches, being feature-matching ones~\cite{SattlerPriorityLoc2017,hfnet} or end-to-end learning-based methods~\cite{dsac*,li2020hscnet,sarlin21pixloc,sc_wls_eccv2022,neumap}, even managing to outperform strong baselines like HLoc~\cite{hfnet} in some cases.
We believe this to be significant for two reasons.
First, \duster{} was never trained for visual localisation in any way.
Second, neither query image nor database images were seen during \duster{}'s training.

\vspace{-1mm}
\subsection{Multi-view Pose Estimation }
\label{ssec:pose_estimation}
We now evaluate \duster{} on multi-view relative pose estimation after the global alignment from \cref{ssec:mvoptim}.

\myParagraph{Datasets.}
Following ~\cite{posediffusion}, we use two multi-view datasets, CO3Dv2~\cite{co3d} and RealEstate10k~\cite{realestate10K} for the evaluation. CO3Dv2 contains 6 million frames extracted from approximately 37k videos, covering 51 MS-COCO categories. The ground-truth camera poses are annotated using COLMAP from 200 frames in each video. RealEstate10k is an indoor/outdoor dataset with 10 million frames from about 80K video clips on YouTube, the camera poses being obtained by SLAM with bundle adjustment. We follow the protocol introduced in~\cite{posediffusion} to evaluate \duster \ on 41 categories from CO3Dv2 and 1.8K video clips from the test set of RealEstate10k. For each sequence, we random select 10 frames and feed all possible 45 pairs to \duster{}. 

\myParagraph{Baselines and metrics.} We compare \duster{} pose estimation results, obtained either from PnP-RANSAC or global alignment, against the learning-based RelPose~\cite{relpose}, PoseReg~\cite{posediffusion} and PoseDiffusion~\cite{posediffusion}, and structure-based PixSFM~\cite{pixsfm}, COLMAP+SPSG (COLMAP~\cite{colmapmvs} extended with SuperPoint~\cite{superpoint} and SuperGlue~\cite{superglue}). 
Similar to~\cite{posediffusion}, we report the Relative Rotation Accuracy (RRA) and Relative Translation Accuracy (RTA) for each image pair to evaluate the relative pose error and select a threshold $\tau=15$ to report RTA@$15$ and RRA$@15$. Additionally, we calculate the mean Average Accuracy (mAA)$@30$, defined as the area under the curve  accuracy of the angular differences at $min({\rm RRA}@30, {\rm RTA}@30)$.

\myParagraph{Results.} 
As shown in Table~\ref{tab:co3d}, \duster{} with global alignment achieves the best overall performance on the two datasets and significantly surpasses the state-of-the-art PoseDiffusion~\cite{posediffusion}. 
Moreover, \duster{} with PnP also demonstrates superior performance over both learning and structure-based existing methods. 
It is worth noting that RealEstate10K results reported for PoseDiffusion are from the model trained on CO3Dv2. 
Nevertheless, we assert that our comparison is justified considering that RealEstate10K is not used either during \duster{}'s training.
We also report performance with less input views (between 3 and 10) in the \supplementary{} (\cref{sec:mvpose}), in which case \duster{} also yields excellent performance on both benchmarks.

\begin{table*}[h!]
\begin{center}
\small
\renewcommand\arraystretch{1.1}
\hspace{-3mm}
\resizebox{0.55\textwidth}{!}{
\begin{tabular}{lccccccccccc}
\hline
\specialrule{1.5pt}{0.5pt}{0.5pt} 
      &        & \multicolumn{4}{c}{Outdoor} & \multicolumn{6}{c}{Indoor} \\
\cline{3-12}
Methods & Train & \multicolumn{2}{c}{DDAD\cite{guizilini203d}} & \multicolumn{2}{c}{KITTI~\cite{geiger13kitti}} & \multicolumn{2}{c}{BONN~\cite{palazzo19refusion}} & \multicolumn{2}{c}{NYUD-v2~\cite{silberman12nyu}} & \multicolumn{2}{c}{TUM~\cite{sturm12benchmark}} \\
\cline{3-12}
 &  & Rel$\downarrow$ & $\delta_{1.25}\uparrow$ & Rel$\downarrow$ & $\delta_{1.25}\uparrow$ & Rel$\downarrow$ & $\delta_{1.25} \uparrow$ & Rel$\downarrow$ & $\delta_{1.25}\uparrow$ & $\operatorname{Rel}\downarrow$ & $\delta_{1.25}\uparrow$ \\
\specialrule{1.5pt}{0.5pt}{0.5pt} 
DPT-BEiT\cite{dpt}                     &D& 10.70&{\bf 84.63}& 9.45&89.27 &  -&-&{\bf 5.40}&{\bf 96.54}&{\bf 10.45}&{\bf 89.68} \\
NeWCRFs\cite{newcrf22}                 &D&{\bf 9.59}&82.92 &{\bf 5.43}&{\bf 91.54}&  -&-&    6.22&95.58 &14.63&82.95 \\
\hline
Monodepth2~\cite{godard19digging}      &SS& 23.91&75.22 &11.42&86.90 &56.49&35.18 &16.19&74.50 &31.20&47.42\\
SC-SfM-Learners~\cite{bian22csdepthv2} &SS& 16.92&77.28 &11.83&86.61 &21.11&71.40 &13.79&79.57 &22.29&64.30 \\
SC-DepthV3~\cite{sun2022scdepthv3}     &SS&{\bf 14.20}&{\bf 81.27}&11.79&86.39 &{\bf 12.58}&{\bf 88.92}&{\bf 12.34}&{\bf 84.80}&{\bf 16.28}&{\bf 79.67}\\
MonoViT\cite{zhao22monovit}            &SS& - & - &{\bf 09.92}&{\bf 90.01}& - & - & - & - & - & \\
\hline
RobustMIX~\cite{ranftl2020robust}   &T& -&- &18.25&76.95 &   -&-     &11.77&90.45 &15.65&{\bf 86.59} \\
SlowTv~\cite{spencer2023kick}        &T&{\bf 12.63}&79.34 &(6.84)&(56.17)&   -&-     &11.59&87.23 &15.02&80.86 \\ %
{\bf {\duster } 224-NoCroCo}        &T& 19.63 & 70.03 & 20.10 & 71.21 & 14.44 & 86.00 & 14.51 & 81.06 & 22.14 & 66.26 \\ 
{\bf {\duster } 224}        &T& 16.32&77.58 &16.97&77.89 &11.05&89.95 &10.28&88.92 &17.61&75.44 \\ 
{\bf {\duster } 512}    &T& 13.88&81.17 &{\bf 10.74}&{\bf 86.60}&{\bf 8.08}&{\bf 93.56}&{\bf  6.50}&94.09&{\bf 14.17}&79.89 \\ %
\specialrule{1.5pt}{0.5pt}{0.5pt} 
\end{tabular}}
\renewcommand\arraystretch{1.2}
\setlength{\tabcolsep}{3pt}
\resizebox{0.45\linewidth}{!}{
\begin{tabular}{lccccc}
\hline
\specialrule{1.5pt}{0.5pt}{0.5pt} 
\multirow{2}{*}{Methods} & \multicolumn{3}{c}{Co3Dv2~\cite{co3d}} &  & RealEstate10K\\ \cline{2-4} \cline{6-6} %
                         & RRA@15  & RTA@15 & mAA(30) &  & mAA(30)       \\ \specialrule{1.5pt}{0.5pt}{0.5pt} 
RelPose~\cite{relpose}               & 57.1    & -      & -       &  & -             \\
Colmap+SPSG~\cite{superpoint, superglue} & 36.1    & 27.3   & 25.3    &  & 45.2          \\
PixSfM~\cite{pixsfm}                 & 33.7    & 32.9   & 30.1    &  & 49.4          \\
PosReg~\cite{posediffusion}          & 53.2    & 49.1   & 45.0    &  & -             \\
PoseDiffusion ~\cite{posediffusion}  & 80.5    & 79.8   & 66.5    &  & 48.0          \\ \hline
\bf{\duster \ 512 (w/ PnP)}               & 94.3    & \bf{88.4}   & \bf{77.2}    &  & 61.2          \\
\bf{\duster \ 512 (w/ GA)}             & \bf{96.2}    & 86.8   & 76.7    &  & \bf{67.7}          \\ \specialrule{1.5pt}{0.5pt}{0.5pt} 
\end{tabular}}
\normalsize
\end{center}
\vspace{-10pt}
\caption{\textbf{Left}: Monocular depth estimation on multiple benchmarks. 
D-Supervised, SS-Self-supervised, T-transfer (zero-shot). 
(Parentheses) refers to training on the same set.
\textbf{Right:} Multi-view pose regression on the CO3Dv2~\cite{co3d} and RealEst10K~\cite{realestate10K} with 10 random frames. 
\label{tab:co3d}
}
\label{tab:duster-mde}
\end{table*}

\begin{table*}
\begin{center}
\renewcommand\arraystretch{1.2}
\setlength{\tabcolsep}{1pt} 
\small
\hspace{-3mm}
\resizebox{\textwidth}{!}{
\begin{tabular}{llccccrrrrrrrrrrrrc}
 \hline
\specialrule{1.5pt}{0.5pt}{0.5pt}
\multicolumn{2}{l}{\multirow{2}{*}{Methods}} & GT & GT & GT & Align & \multicolumn{2}{c}{KITTI} & \multicolumn{2}{c}{ScanNet} & \multicolumn{2}{c}{ETH3D} & \multicolumn{2}{c}{DTU} & \multicolumn{2}{c}{T\&T} & \multicolumn{3}{c}{Average} \\
\cline{7-19}
 && Pose & Range & Intrinsics &  & rel $\downarrow$ & $\tau \uparrow$ & rel $\downarrow$ & $\tau \uparrow$ & rel $\downarrow$ & $\tau \uparrow$ & rel $\downarrow$ & $\tau \uparrow$ & rel$\downarrow$ & $\tau \uparrow$ & rel$\downarrow$ & $\tau \uparrow$ & time (s)$\downarrow$ \\
\specialrule{1.5pt}{0.5pt}{0.5pt}
\multirow{2}{*}{(a)} & COLMAP~\cite{colmapsfm,colmapmvs} & $\checkmark$ & $\times$ & $\checkmark$ & $\times$ &{\bf 12.0}&{\bf 58.2} &{\bf 14.6}&{\bf 34.2}&{\bf 16.4}&{\bf 55.1}&{\bf 0.7}&{\bf 96.5}&{\bf 2.7}& 95.0 &{\bf 9.3} & {\bf 67.8} & $\approx$ 3 min  \\
&COLMAP Dense~\cite{colmapsfm,colmapmvs} & $\checkmark$&$\times$ & $\checkmark$ & $\times$ & 26.9 & 52.7 & 38.0 & 22.5 & 89.8 & 23.2 & 20.8 & 69.3 & 25.7 & 76.4 & 40.2 & 48.8 & $\approx$ 3 min \\
\hline
\multirow{5}{*}{(b)} & MVSNet~\cite{mvsnet} & $\checkmark$ & $\checkmark$ &$\checkmark$ & $\times$ & 22.7 & 36.1 & 24.6 & 20.4 & 35.4 & 31.4 & (1.8) & $(86.0)$ & 8.3 & 73.0 & 18.6 & 49.4 & 0.07 \\
& MVSNet Inv. Depth~\cite{mvsnet} & $\checkmark$ & $\checkmark$ &$\checkmark$ & $\times$ & 18.6 & 30.7 & 22.7 & 20.9 & 21.6 & 35.6 & (1.8) & $(86.7)$ & 6.5 & 74.6 & 14.2 & 49.7 & 0.32\\
& Vis-MVSSNet~\cite{vis-mvsnet} & $\checkmark$ & $\checkmark$ & $\checkmark$ & $\times$ &{\bf 9.5}&{\bf 55.4}& 8.9 & 33.5 &{\bf 10.8}&{\bf 43.3}&{\bf (1.8)} &{\bf (87.4)} &{\bf 4.1}&{\bf 87.2} &{\bf 7.0} &{\bf 61.4} & 0.70 \\
& MVS2D ScanNet~\cite{MVS2D} & $\checkmark$ & $\checkmark$ & $\checkmark$ & $\times$ & 21.2 & 8.7 & (27.2) & (5.3) & 27.4 & 4.8 & 17.2 & 9.8 & 29.2 & 4.4 & 24.4 & 6.6 & \textbf{0.04} \\
& MVS2D DTU~\cite{MVS2D} & $\checkmark$ & $\checkmark$& $\checkmark$ & $\times$ & 226.6 & 0.7 & 32.3 & 11.1 & 99.0 & 11.6 & (3.6) & (64.2) & 25.8 & 28.0 & 77.5 & 23.1 & 0.05  \\
\hline
\multirow{8}{*}{(c)} & DeMon~\cite{demon} & $\checkmark$ & $\times$ &$\checkmark$ & $\times$ & 16.7 & 13.4 & 75.0 & 0.0 & 19.0 & 16.2 & 23.7 & 11.5 & 17.6 & 18.3 & 30.4 & 11.9 & 0.08  \\
& DeepV2D KITTI~\cite{deepv2d} & $\checkmark$ & $\times$ &$\checkmark$ & $\times$ & (20.4) & (16.3) & 25.8 & 8.1 & 30.1 & 9.4 & 24.6 & 8.2 & 38.5 & 9.6 & 27.9 & 10.3  & 1.43\\
& DeepV2D ScanNet~\cite{deepv2d} & $\checkmark$ & $\times$ &$\checkmark$ & $\times$ & 61.9 & 5.2 & (3.8) & (60.2) & 18.7 & 28.7 & 9.2 & 27.4 & 33.5 & 38.0 & 25.4 & 31.9  & 2.15 \\
& MVSNet~\cite{mvsnet} & $\checkmark$ & $\times$ &$\checkmark$ & $\times$ & 14.0 & 35.8 & 1568.0 & 5.7 & 507.7 & 8.3 & (4429.1) & (0.1) & 118.2 & 50.7 & 1327.4 & 20.1 & 0.15 \\
& MVSNet Inv. Depth~\cite{mvsnet} & $\checkmark$ & $\times$ &$\checkmark$ & $\times$ & 29.6 & 8.1 & 65.2 & 28.5 & 60.3 & 5.8 & (28.7) & (48.9) & 51.4 & 14.6 & 47.0 & 21.2 & 0.28  \\
& Vis-MVSNet \cite{vis-mvsnet} & $\checkmark$ & $\times$ &$\checkmark$ & $\times$ & 10.3 &{\bf 54.4}& 84.9 & 15.6 & 51.5 & 17.4 & (374.2) & (1.7) & 21.1 & 65.6 & 108.4 & 31.0 &0.82 \\
& MVS2D ScanNet~\cite{MVS2D} & $\checkmark$ &$\times$ &$\checkmark$ &$\times$ & 73.4 & 0.0 & (4.5) & (54.1) & 30.7 & 14.4 & 5.0 & 57.9 & 56.4 & 11.1 & 34.0 & 27.5 & \textbf{0.05}  \\
& MVS2D DTU~\cite{MVS2D} &$\checkmark$ & $\times$ & $\checkmark$ & $\times$ & 93.3 & 0.0 & 51.5 & 1.6 & 78.0 & 0.0 & (1.6) & (92.3) & 87.5 & 0.0 & 62.4 & 18.8 & 0.06 \\
& Robust MVD Baseline~\cite{robust_mvd} &$\checkmark$&$\times$&$\checkmark$&$\times$ &{\bf 7.1} & 41.9 & {\bf 7.4}&{\bf 38.4}&{\bf 9.0}&{\bf 42.6}&{\bf 2.7}&{\bf 82.0}&{\bf 5.0}&{\bf 75.1}&{\bf 6.3} &{\bf 56.0} & 0.06\\
\hline
\multirow{7}{*}{(d)} & DeMoN~\cite{demon} &$\times$&$\times$ &$\checkmark$& $\|\mathbf{t}\|$ & 15.5 & 15.2 & 12.0 & 21.0 & 17.4 & 15.4 & 21.8 & 16.6 & 13.0 & 23.2 & 16.0 & 18.3 & 0.08 \\
& DeepV2D KITTI~\cite{deepv2d} &$\times$&$\times$&$\checkmark$&med& (3.1) & (74.9) & 23.7 & 11.1 & 27.1 & 10.1 & 24.8 & 8.1 & 34.1 & 9.1 & 22.6 & 22.7 & 2.07 \\
& DeepV2D ScanNet~\cite{deepv2d} &$\times$&$\times$&$\checkmark$& med &10.0 & 36.2 &\bf{(4.4)} & (54.8) & 11.8 & 29.3 & 7.7 & 33.0 & 8.9 & 46.4 & 8.6 & 39.9 & 3.57  \\
& \bf{{\duster} 224-NoCroCo} &$\times$&$\times$&$\times$&med& 15.14&21.16 & 7.54&40.00 & 9.51&40.07 & 3.56&62.83 & 11.12&37.90 & 9.37 & 40.39 & \textbf{0.05} \\
& \bf{{\duster} 224}     &$\times$&$\times$&$\times$&med& 15.39&26.69 & (5.86)&(50.84) & 4.71&61.74 &{\bf 2.76}&{\bf 77.32}& 5.54&56.38 & 6.85&54.59 & \textbf{0.05} \\
& \bf{{\duster} 512} &$\times$&$\times$&$\times$&med&{\bf 9.11}&{\bf 39.49}& (4.93)&({\bf 60.20}) &{\bf 2.91}&{\bf 76.91}& 3.52&69.33 &{\bf 3.17}&{\bf 76.68}&{\bf 4.73}&{\bf 64.52} & 0.13\\
\specialrule{1.5pt}{0.5pt}{0.5pt}
\end{tabular}}

\normalsize
\caption{
\textbf{Multi-view depth evaluation} with different settings: 
a) Classical approaches; 
b) with poses and depth range, without alignment;
c) absolute scale evaluation with poses, without depth range and alignment; 
d) without poses and depth range, but with alignment.
(Parentheses) denote training on data from the same domain. The best results for each setting are in \textbf{bold}.
}
\label{tab:mvd}
\end{center}
\vspace*{-1mm}
\end{table*}

\vspace{-2mm}
\subsection{Monocular Depth}
For this monocular task, we simply feed the same input image $I$ to the network as $\F(I,I)$. 
By design, depth prediction is simply the $z$ coordinate in the predicted 3D pointmap.

\label{ssec:duster_mde}
\myParagraph{Datasets and metrics.}
We benchmark \duster{} on two outdoor (DDAD~\cite{guizilini203d}, KITTI~\cite{geiger13kitti}) and three indoor (NYUv2~\cite{silberman12nyu}, BONN~\cite{palazzo19refusion}, TUM~\cite{sturm12benchmark}) datasets. 
We compare \duster{ }'s performance to state-in-the-art methods categorized in supervised, self-supervised and zero-shot settings, this last category corresponding to \duster{}.  
We use two metrics commonly used in the monocular depth evaluations~\cite{bian22csdepthv2,spencer2023kick}: the absolute relative error $AbsRel$ between target $y$ and prediction $\hat y$, $AbsRel = |y - \hat{y}|/y$, and the prediction threshold accuracy, $\delta_{1.25} = \max ({\hat y}/y, y/ {\hat y}) < 1.25$.

\myParagraph{Results.}
In zero-shot setting, the state of the art is represented by the recent SlowTv~\cite{spencer2023kick}.
This approach collected a large mixture of curated datasets with urban, natural, synthetic and indoor scenes, and trained one common model. 
For every dataset in the mixture, camera parameters are known or estimated with COLMAP. 
As Table~\ref{tab:duster-mde} shows, {\duster } adapts well to outdoor and indoor environments. It outperforms the self-supervised baselines~\cite{godard19digging,sun2022scdepthv3,bian22csdepthv2} and performs on-par with state-of-the-art supervised baselines~\cite{newcrf22,dpt}.

\subsection{Multi-view Depth} 
\label{ssec:duster_mvd}

We evaluate \duster{} for the task of multi-view stereo depth estimation. 
Likewise, we extract depthmaps as the $z$-coordinate of predicted pointmaps. 
In the case where multiple depthmaps are available for the same image, we rescale all predictions to align them together and aggregate all predictions via a simple averaging weighted by the confidence.

\myParagraph{Datasets and metrics.}
Following~\cite{robust_mvd}, we evaluate it on the DTU~\cite{dtu}, ETH3D~\cite{eth3d}, Tanks and Temples~\cite{tandt}, and ScanNet~\cite{scannet} datasets. 
We report the Absolute Relative Error (rel) and Inlier Ratio $(\tau)$ with a threshold of 1.03 on each test set and the averages across all test sets. Note that we do not leverage the \emph{ground-truth} camera parameters and poses nor the \emph{ground-truth} depth ranges, so our predictions are only valid up to a scale factor. 
In order to perform quantitative measurements, we thus normalize  predictions using the medians of the predicted depths and the ground-truth ones, as advocated in~\cite{robust_mvd}. 

\myParagraph{Results.}
We observe in Table~\ref{tab:mvd} that \duster{} achieves state-of-the-art accuracy on ETH-3D and outperforms most recent state-of-the-art methods overall, even those using ground-truth camera poses.
Timewise, our approach is also much faster than the traditional COLMAP pipeline~\cite{colmapsfm,colmapmvs}.
This showcases the applicability of our method on a large variety of domains, either indoors, outdoors, small scale or large scale scenes, while not having been trained on the test domains, except for the ScanNet test set, since the train split is part of the Habitat dataset.

\subsection{3D Reconstruction}
\label{ssec:duster_mvs}

Finally, we measure the quality of our full reconstructions obtained after the global alignment procedure described in~\cref{ssec:mvoptim}. 
We again emphasize that our method is the first one to enable global unconstrained MVS, %
in the sense that we have no prior knowledge regarding the camera intrinsic and extrinsic parameters. In order to quantify the quality of our reconstructions, we simply align the predictions to the ground-truth coordinate system. This is done by fixing the parameters as constants in~\cref{eq:pose_optim}. This leads to consistent 3D reconstructions expressed in the coordinate system of the ground-truth. 

\myParagraph{Datasets and metrics.}
We evaluate our predictions on the DTU~\cite{dtu} dataset. 
We apply our network in a zero-shot setting, \ie we do not finetune on the DTU train set and apply our model as is.
In \cref{tab:mvs_dtu} we report the averaged accuracy, averaged completeness and overall averaged error metrics as provided by the authors of the benchmarks. 
The accuracy for a point of the reconstructed shape is defined as the smallest Euclidean distance to the ground-truth, and
the completeness of a point of the ground-truth as the
smallest Euclidean distance to the reconstructed shape. 
The overall is simply the mean of both previous metrics. 

\myParagraph{Results.}
Our method does not reach the accuracy levels of the best methods.
In our defense, these methods all leverage GT poses and train specifically on the DTU train set whenever applicable. 
Furthermore, best results on this task are usually obtained via sub-pixel accurate triangulation, requiring the use of explicit camera parameters, whereas our approach relies on regression, which is known to be less accurate. 
Yet, without prior knowledge about the cameras, we reach an average accuracy of $2.7mm$, with a completeness of $0.8mm$, for an overall average distance of $1.7mm$. 
We believe this level of accuracy to be of great use in practice, considering the \emph{plug-and-play} nature of our approach.

\subsection{Ablations}
\label{ssec:ablations}

We ablate the impact of the CroCo pretraining and image resolution on \duster{}'s performance.
We report results in tables \cref{tab:cambridge}, \cref{tab:co3d}, \cref{tab:mvd} for the tasks mentioned above.
Overall, the observed consistent improvements suggest the crucial role of pretraining and high resolution in modern data-driven approaches, as also noted by~\cite{croco_v2, dino_v2}.

\begin{table} 
\begin{center}

\renewcommand\arraystretch{1.2}
\setlength{\tabcolsep}{3pt}
\resizebox{\linewidth}{!}{
\begin{tabular}{llcccc}
\specialrule{1.5pt}{0.5pt}{0.5pt}
& Methods & GT cams & Acc.$\downarrow$ & Comp.$\downarrow$ & Overall$\downarrow$       \\
\hline
\multirow{4}{*}{(a)} & Camp~\cite{camp} & $\checkmark$ & 0.835 & 0.554 & 0.695 \\
&Furu~\cite{furu} & $\checkmark$ & 0.613 & 0.941 & 0.777 \\
&Tola~\cite{tola} & $\checkmark$ & 0.342 & 1.190 & 0.766 \\
&Gipuma~\cite{gipuma} & $\checkmark$ &\textbf{ 0.283} & 0.873 & 0.578\\
\specialrule{1.5pt}{0.5pt}{0.5pt}
\multirow{4}{*}{(b)} &MVSNet~\cite{mvsnet} & $\checkmark$ &0.396 & 0.527 & 0.462 \\
&CVP-MVSNet~\cite{cvp-mvsnet} & $\checkmark$ & 0.296 & 0.406 & 0.351 \\
&UCS-Net~\cite{ucs-net} & $\checkmark$ & 0.338 & 0.349 & 0.344 \\
& CER-MVS~\cite{cermvs} & $\checkmark$ & 0.359 & 0.305 & 0.332 \\
& CIDER~\cite{cider} & $\checkmark$ & 0.417 & 0.437 & 0.427 \\
& CasMVSNet~\cite{casmvsnet} & $\checkmark$ & 0.325 & 0.385 & 0.355 \\
& PatchmatchNet~\cite{pathcmatchnet} & $\checkmark$ & 0.427 & 0.277 & 0.352 \\
& GeoMVSNet~\cite{geomvsnet} & $\checkmark$ & 0.331 & \textbf{0.259} & \textbf{0.295} \\
\specialrule{1.5pt}{0.5pt}{0.5pt}
&{\bf \duster \ 512} & $\times$ &   2.677  &  0.805  & 1.741  \\ %
\specialrule{1.5pt}{0.5pt}{0.5pt}
\end{tabular}
}
\normalsize
\caption{
\textbf{MVS results} on the DTU dataset, in \emph{mm}. Traditional handcrafted methods (a) have been overcome by learning-based approaches (b) that train on this specific domain. 
\label{tab:mvs_dtu}
}
\end{center}
\end{table}

\section{Conclusion}
We presented a novel paradigm to solve not only 3D reconstruction in-the-wild without prior information about scene nor cameras, but a whole variety of 3D vision tasks as well.

\clearpage
\newcounter{myc}[part] 
\renewcommand{\thesection}{\Alph{myc}} %
\let\osection\section %
\renewenvironment{section}{\stepcounter{myc}\osection} 
\maketitlesupplementary
\begin{figure}
    \centering
    \includegraphics[width=\linewidth]{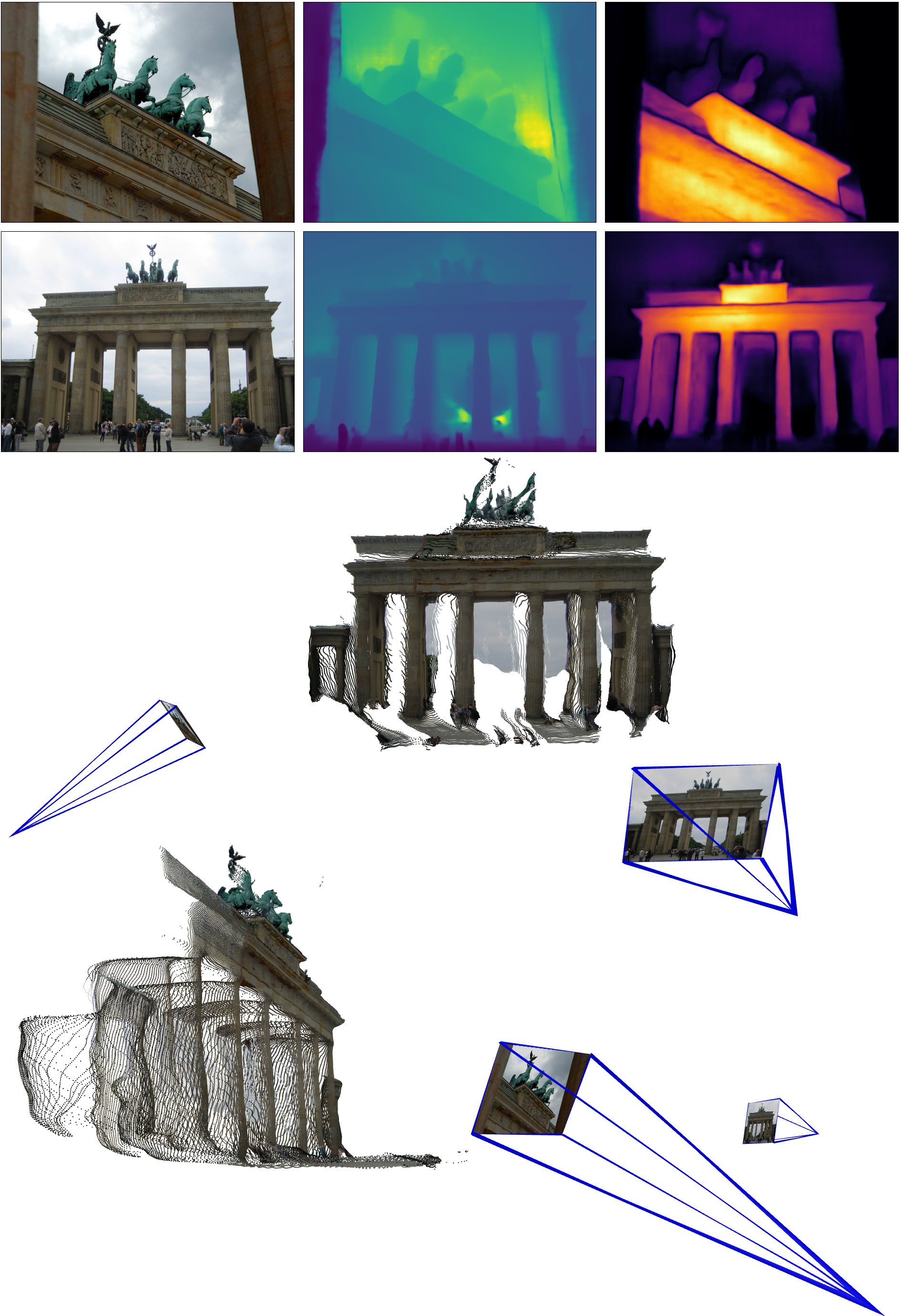}
    \caption{\textbf{Example of 3D reconstruction} of an unseen MegaDepth scene from two images (top-left).
    Note this is the \textbf{raw output} of the network, \ie we show the output depthmaps (top-center, see \cref{eq:depthmap}) and confidence maps (top-right), as well as two different viewpoints on the colored pointcloud (middle and bottom). 
    Camera parameters are recovered from the raw pointmaps, see~\cref{ssec:postproc} in the main paper.
    \duster{} handles strong viewpoint and focal changes without apparent problems}
    \label{fig:megadepth_1}
\end{figure}

\begin{figure}
    \centering
    \includegraphics[width=\linewidth]{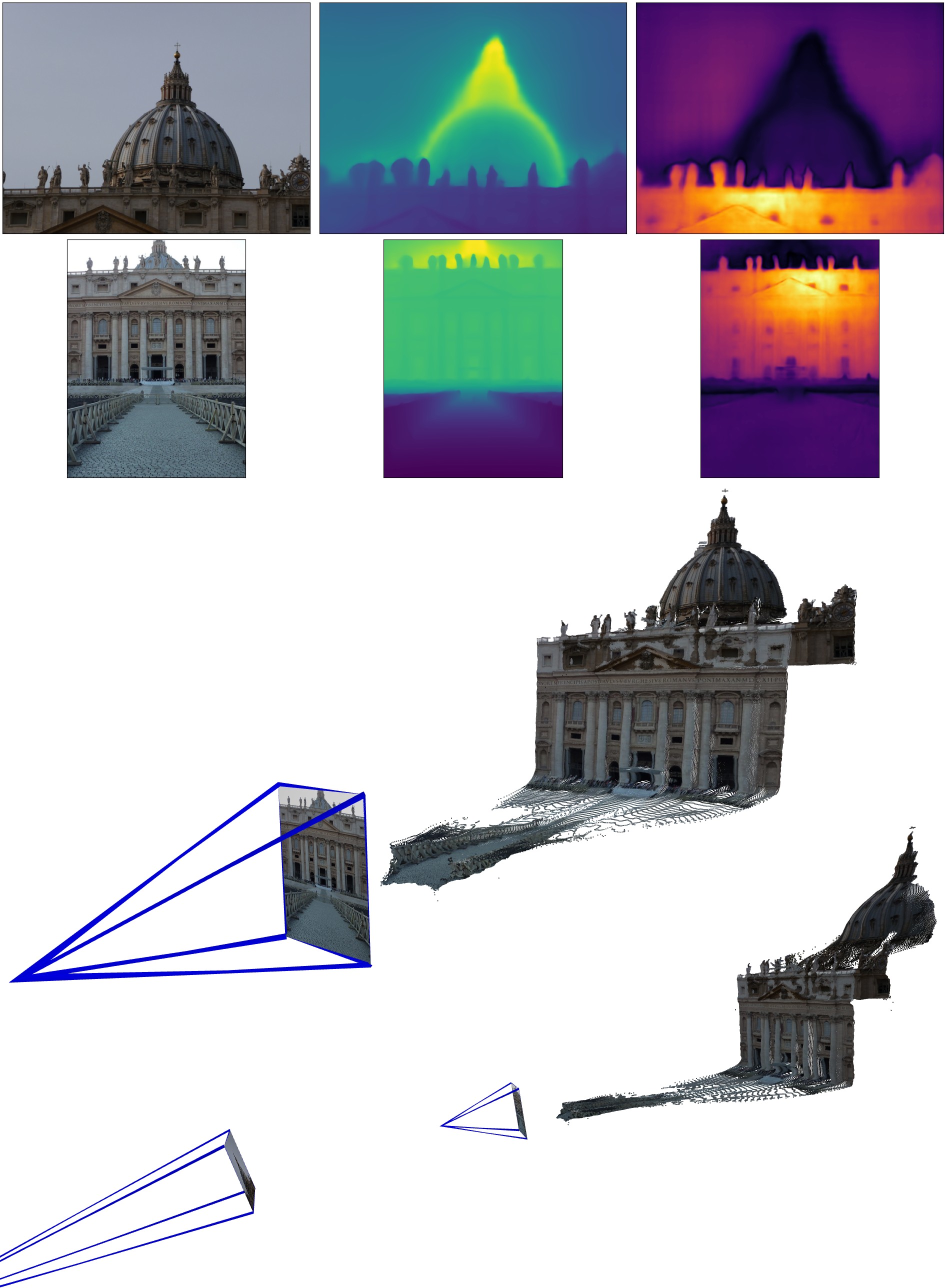}    
    \caption{\textbf{Example of 3D reconstruction} of an unseen MegaDepth~\cite{megadepth} scene from two images only.
    Note this is the \textbf{raw output} of the network, \ie we show the output depthmaps (top-center) and confidence maps (top-right), as well as different viewpoints on the colored pointcloud (middle and bottom). 
    Camera parameters are recovered from the raw pointmaps, see~\cref{ssec:postproc} in the main paper.
    \duster{} handles strong viewpoint and focal changes without apparent problems}
    \label{fig:megadepth_2}
\end{figure}

\begin{figure*}
    \centering
    \includegraphics[width=0.9\linewidth]{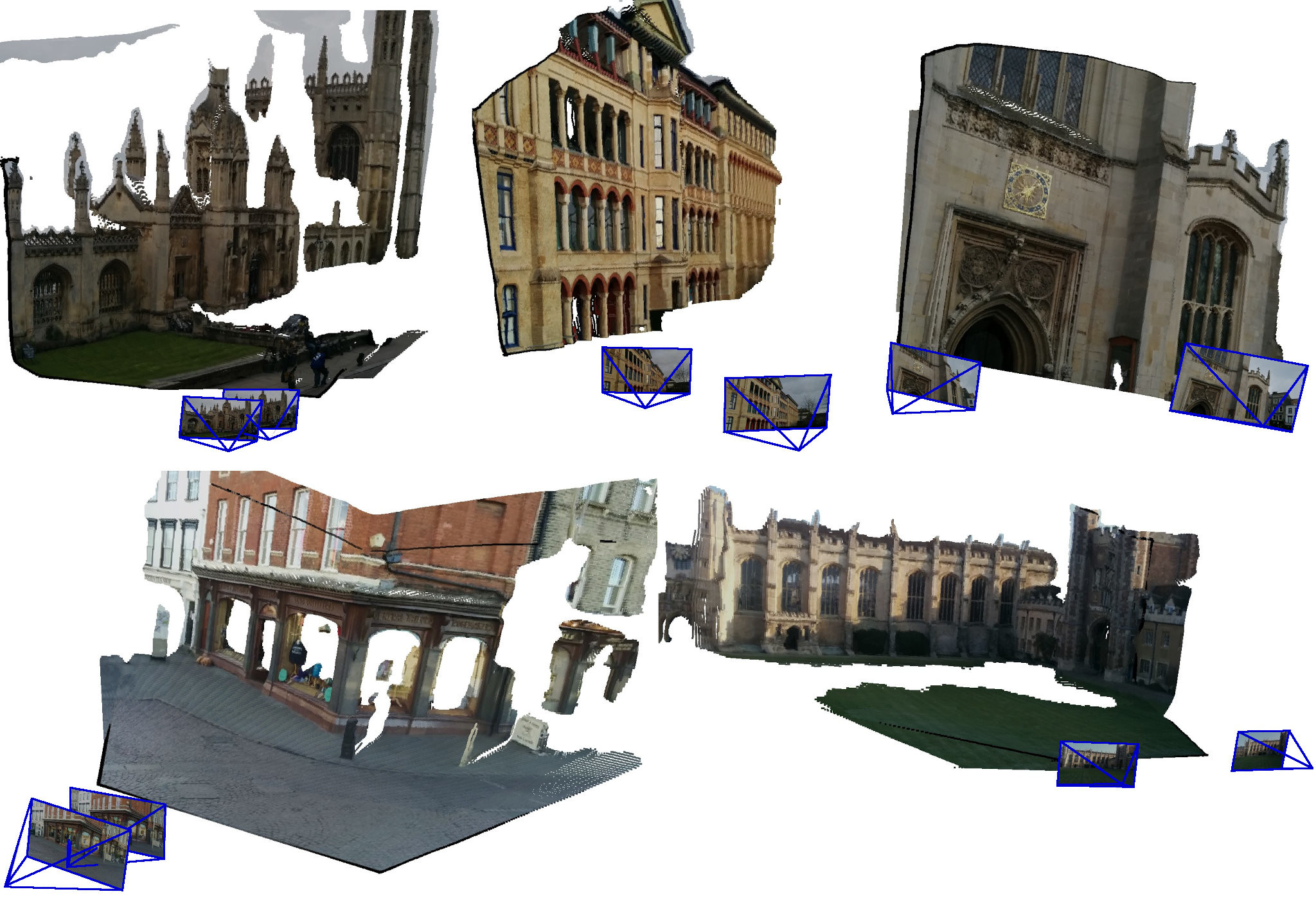}
    \caption{\textbf{Example of 3D reconstruction} from two images only of unseen scenes, namely KingsCollege(Top-Left), OldHospital (Top-Middle), StMarysChurch(Top-Right), ShopFacade(Bottom-Left), GreatCourt(Bottom-Right).
    Note this is the \textbf{raw output} of the network, \ie we show new viewpoints on the colored pointclouds. 
    Camera parameters are recovered from the raw pointmaps, see~\cref{ssec:postproc} in the main paper.}
    \label{fig:cambridge}
\end{figure*}

\begin{figure*}
    \centering
    \includegraphics[width=\linewidth]{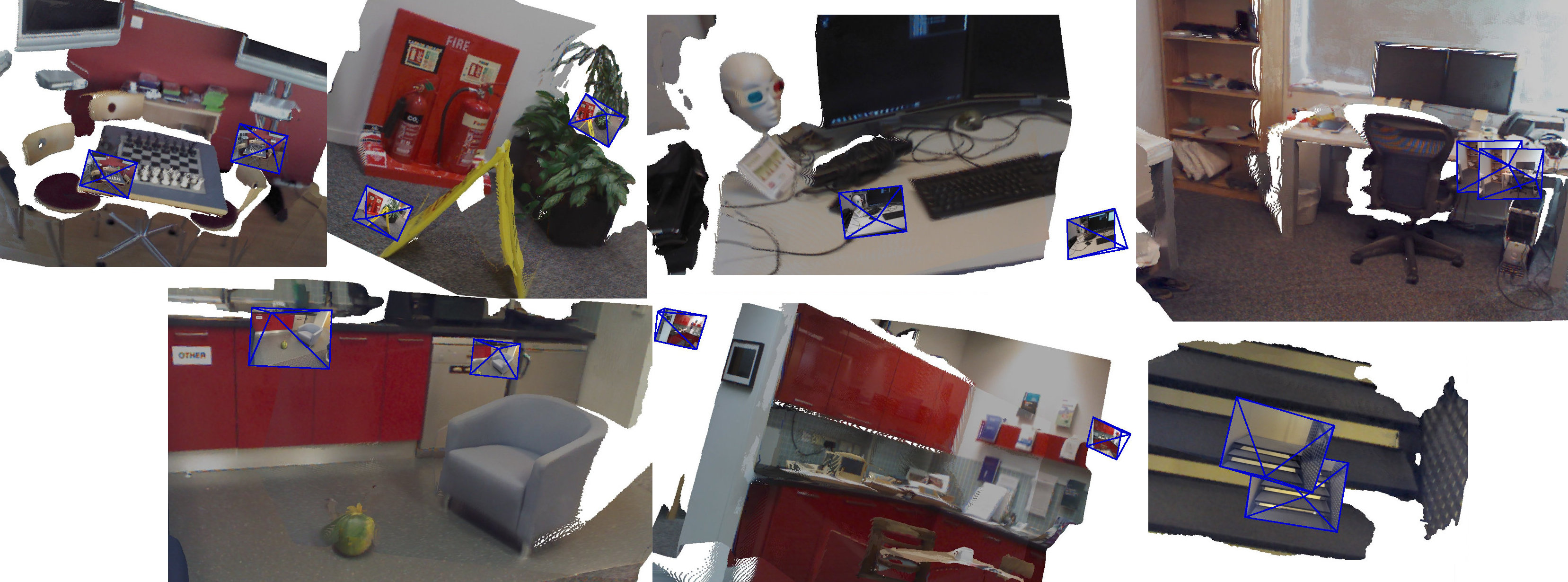}
    \caption{\textbf{Example of 3D reconstruction} from two images only of unseen scenes, namely Chess, Fire, Heads, Office (Top-Row), Pumpkin, Kitchen, Stairs (Bottom-Row).
    Note this is the \textbf{raw output} of the network, \ie we show new viewpoints on the colored pointclouds. 
    Camera parameters are recovered from the raw pointmaps, see~\cref{ssec:postproc} in the main paper.}
    \label{fig:7scenes}
\end{figure*}

\begin{figure*}
    \centering
    $\vcenter{\hbox{
    \resizebox{0.49\linewidth}{!}{
    \begin{tabular}{@{}l@{}}
    \includegraphics[width=1\linewidth]{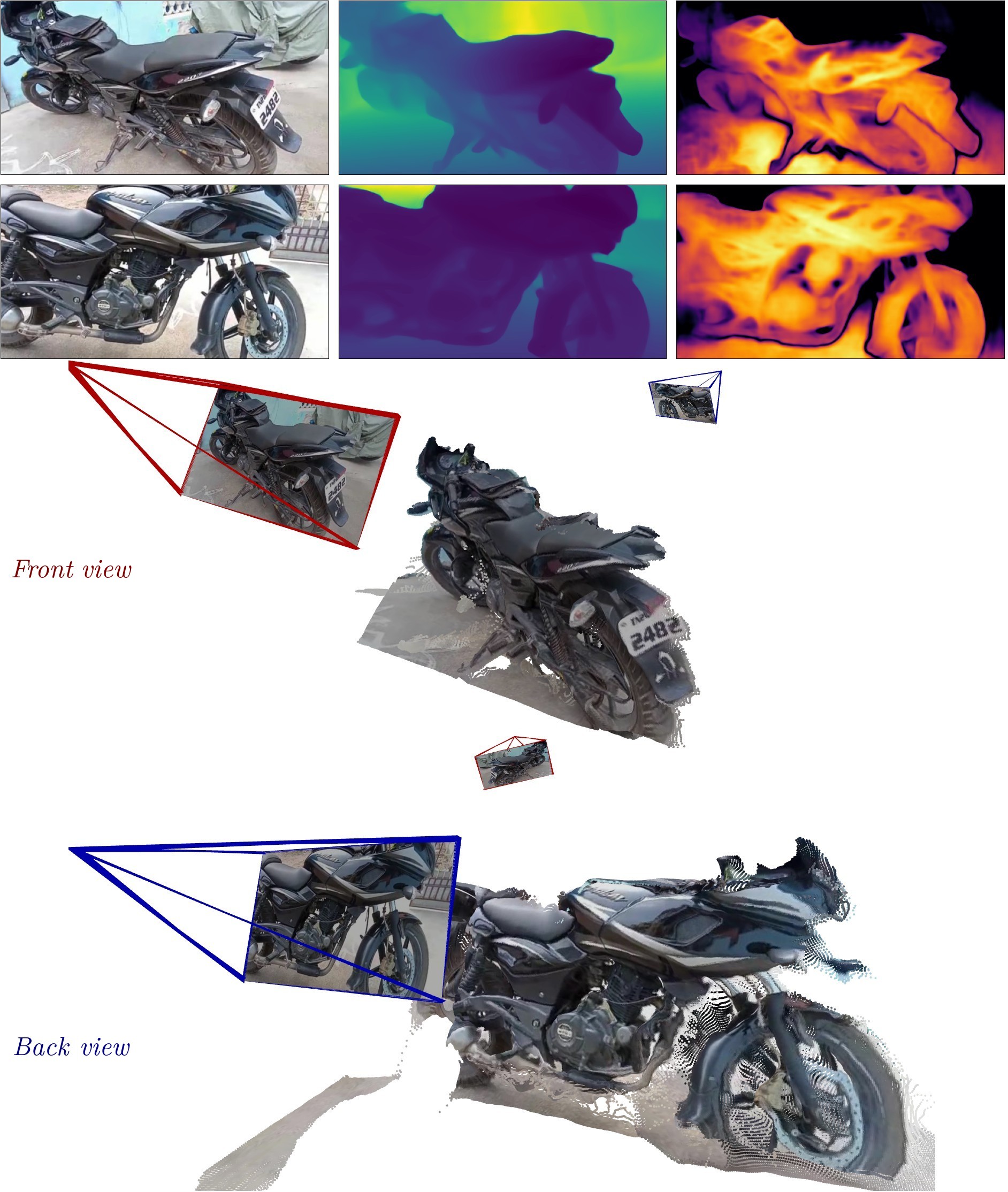}\\[2cm]
    \includegraphics[width=1\linewidth]{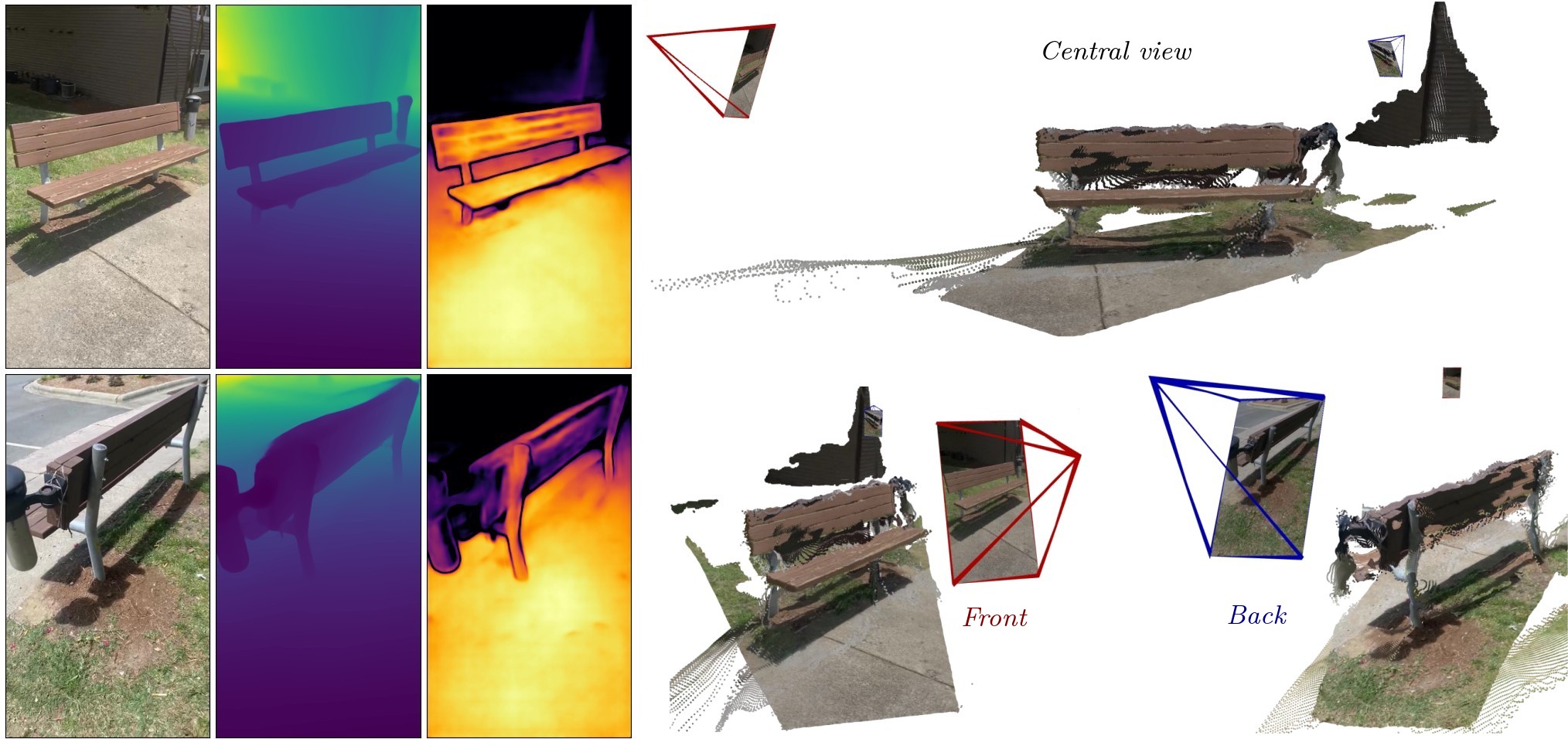}\\[2cm]
    \includegraphics[width=1\linewidth]{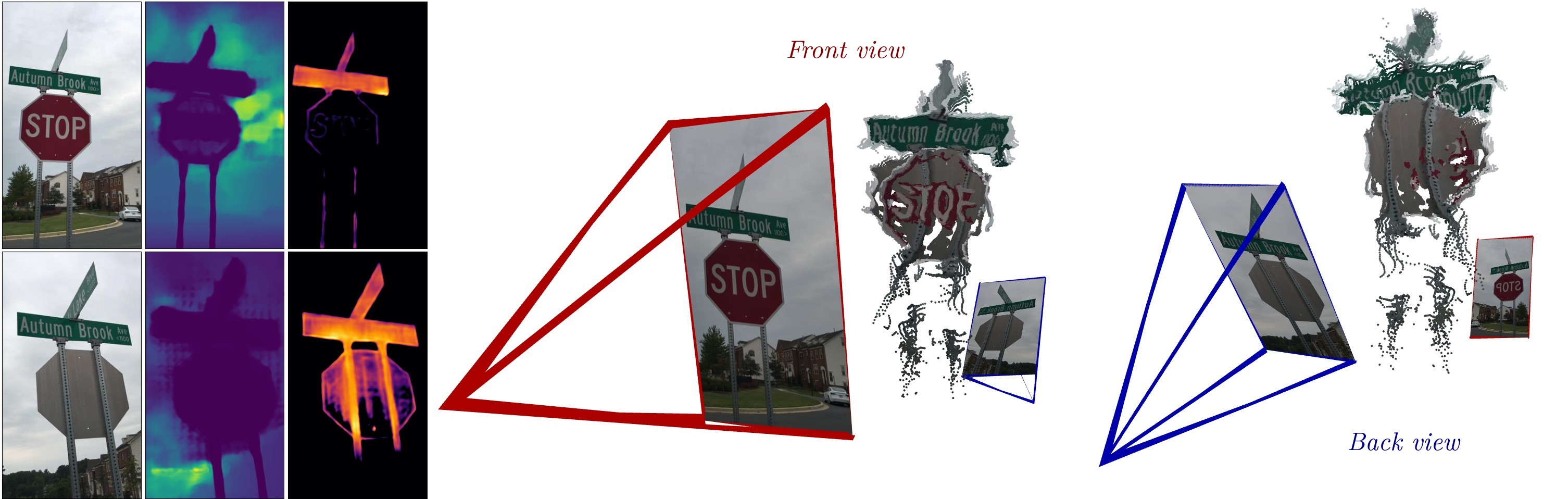}
    \end{tabular}
    }}}$
    $\vcenter{\hbox{
    \includegraphics[width=0.49\linewidth]{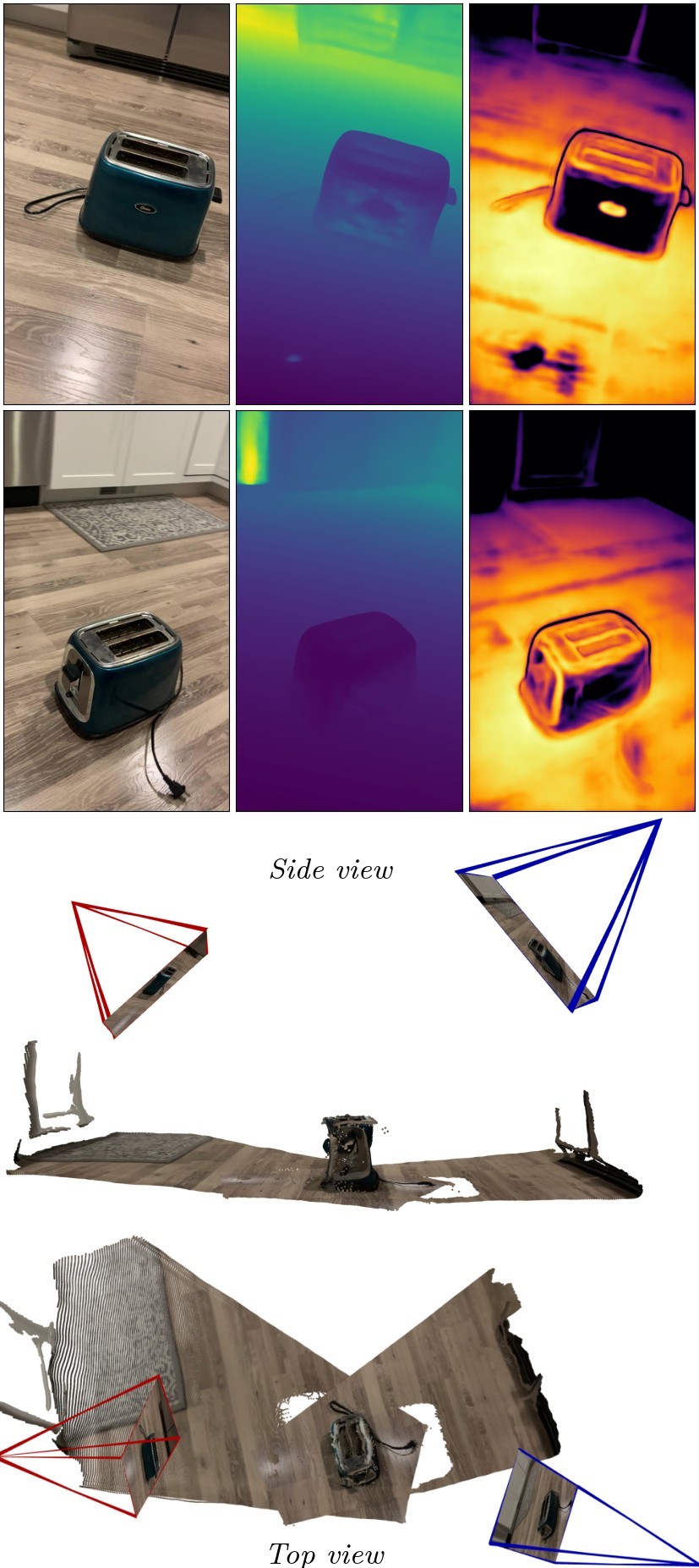}
    }}$
    \caption{\textbf{Examples of 3D reconstructions from nearly opposite viewpoints}.
        For each of the 4 cases (motorcycle, toaster, bench, stop sign), we show the two input images (top-left) and the \textbf{raw output} of the network: output depthmaps (top-center) and confidence maps (top-right), as well as two different views on the colored point-clouds (middle and bottom). 
        Camera parameters are recovered from the raw pointmaps, see~\cref{ssec:postproc} in the main paper.
        \duster{} handles drastic viewpoint changes without apparent issues, even when there is almost no overlapping visual content between images, \eg for the stop sign and motorcycle.
        Note that these example cases are \emph{not} cherry-picked; they are randomly chosen from the set of unseen CO3D\_v2 sequences. Please refer to the \href{https://dust3r.europe.naverlabs.com}{video} for animated visualizations.
    }
    \label{fig:opposite}
\end{figure*}

\begin{figure*}
    \centering
    \includegraphics[width=0.8\linewidth]{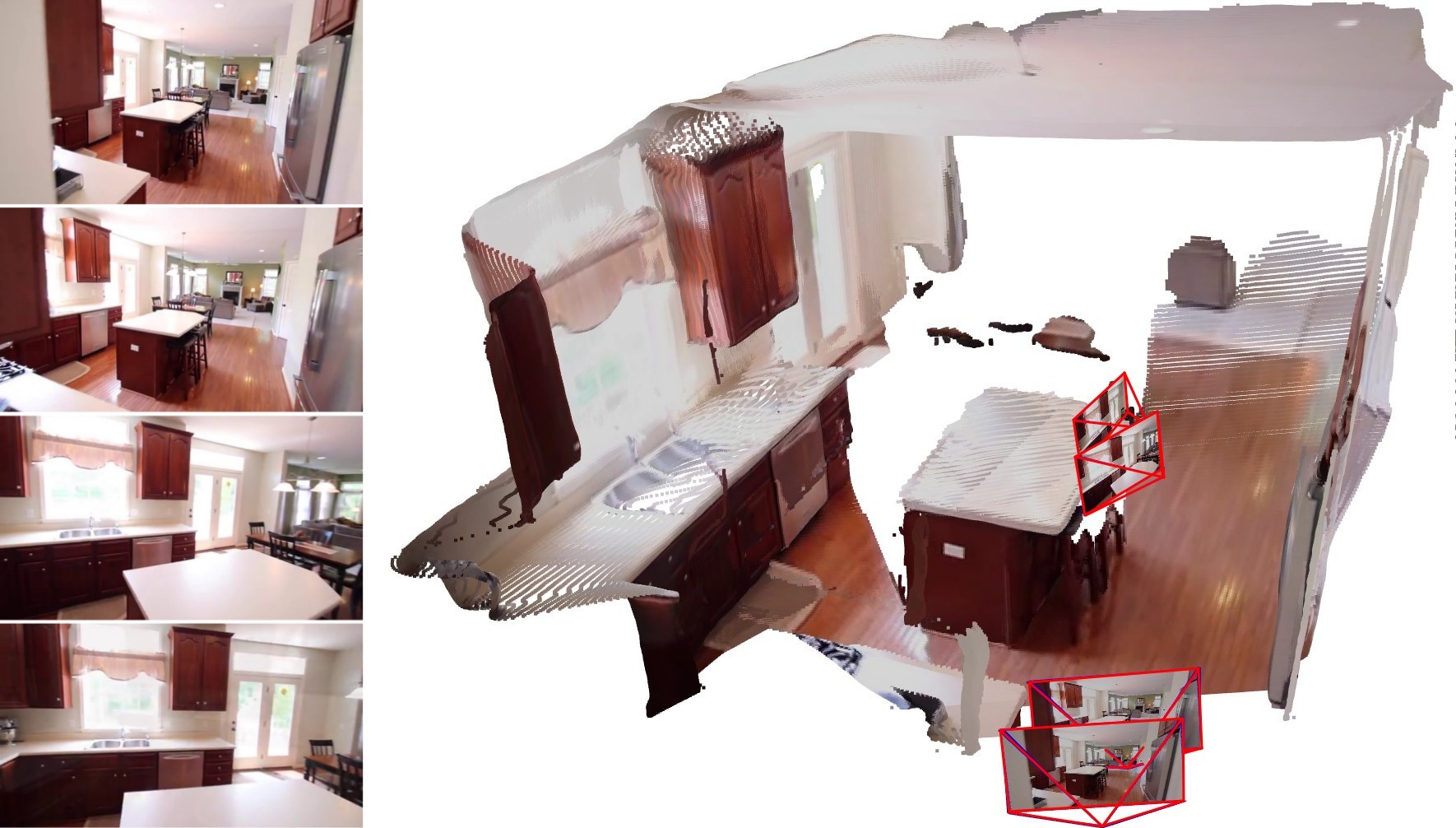}
    \caption{Reconstruction example from 4 random frames of a RealEstate10K indoor sequence, after global alignment.
    On the left-hand side, we show the 4 input frames, and on the right-hand side the resulting point-cloud and the recovered camera intrinsics and poses.}
    \label{fig:realestate}
\end{figure*}

\clearpage
\section*{Appendix}

This \supplementary{} provides additional details and qualitative results of \duster{}. We first present in~\cref{sec:quali} qualitative pairwise predictions of the presented architecture on challenging real-life datasets. This section also contains the description of the video accompanying this material. We then propose an extended related works in~\cref{sec:suprel}, encompassing a wider range of methodological families and geometric vision tasks. \cref{sec:mvpose} provides auxiliary ablative results on multi-view pose estimation, that did not fit in the main paper. We then report in~\cref{sec:visloc} results on an experimental visual localization task, where the camera intrinsics are unknown. 
Finally, we details the training and data augmentation procedures in~\cref{sec:training_details}.

\section{Qualitative results}
\label{sec:quali}

\paragraph{Point-cloud visualizations.}
We present some visualization of \duster{}'s pairwise results in \cref{fig:megadepth_1,fig:megadepth_2,fig:cambridge,fig:7scenes,fig:opposite}.
Note these scenes were never seen during training and were not cherry-picked.
Also, we did not post-process these results, except for filtering out low-confidence points (based on the output confidence) and removing sky regions for the sake of visualization, \ie these figures accurately represent the raw output of \duster{}.
Overall, the proposed network is able to perform highly accurate 3D reconstruction from just two images.
In \cref{fig:realestate}, we show the output of \duster{} after the global alignment stage. In this case, the network has processed all pairs of the 4 input images, and outputs 4 spatially consistent pointmaps along with the corresponding camera parameters.

Note that, for the case of image sequences captured with the same camera, we never enforce the fact that camera intrinsics must be identical for every frame, \ie all intrinsic parameters are optimized independently.
This remains true for all results reported in this \supplementary{} and in the main paper, \eg on multi-view pose estimation with the CO3Dv2~\cite{co3d} and RealEstate10K~\cite{realestate10K} datasets.

\myParagraph{Supplementary Video.}
We attach to this \supplementary{} a video showcasing the different steps of \duster{}.
In the video, we demonstrate dense 3D reconstruction from a small set of raw RGB images, without using any ground-truth camera parameters (\ie unknown intrinsic and extrinsic parameters). 
We show that our method can seamlessly handle monocular predictions, and is able to perform reconstruction and camera pose estimation in extreme binocular cases, where the cameras are facing each other. In addition, we show some qualitative reconstructions of rather large scale scenes from the ETH3D dataset~\cite{eth3d}.

\section{Extended Related Work}
\label{sec:suprel}

For the sake of exposition, Section 2 of the main paper covered only some (but not all) of the most related works. Because this work covers a large variety of geometric tasks, we complete it in this section with a few equally important topics. 

\myParagraph{Implicit Camera Models.}
In our work, we do not explicitly output camera parameters.
Likewise, there are several works aiming to express 3D shapes in a canonical space that is not directly related to the input viewpoint. 
Shapes can be stored as occupancy in regular grids~\cite{3dr2n2, divandimp, pix2vox, pix2vox++,matryoshkanet,wang21multiview,shi21_3dretr}, octree structures~\cite{ogn}, collections of parametric surface elements~\cite{atlasnet}, point clouds encoders~\cite{densepcr,3dlmnet,psgn}, free-form deformation of template meshes~\cite{image2mesh} or per-view depthmaps~\cite{li18}. 
While these approaches arguably perform classification and not actual 3D reconstruction~\cite{tatarchenko19}, all-in-all, they work only in very constrained setups, usually on ShapeNet~\cite{shapenet} and have trouble generalizing to natural scenes with non object-centric views~\cite{photoba}. The question of how to express a complex scene with several object instances in a single canonical frame had yet to be answered: in this work, we also express the reconstruction in a canonical reference frame, but thanks to our scene representation (pointmaps), we still preserve a relationship between image pixels and the 3D space, and we are thus able to perform 3D reconstruction consistently.

\myParagraph{Dense Visual SLAM.} 
In visual SLAM, early works on dense 3D reconstruction and ego-motion estimation utilized active depth sensors~\cite{newcombe11dtam,deeptam,demon}.
Recent works on dense visual SLAM from RGB video stream are able to produce high-quality depth maps and camera trajectories 
\cite{bloesch18codeslam, czarnowski20-deepfactors, Sun_2020_CVPR, cnn-slam, teed21droid-slam, smith2023flowcam}, but they inherit the traditional limitations of SLAM, \eg noisy predictions, drifts and outliers in the pixel correspondences. 
To make the 3D reconstruction more robust, R3D3~\cite{r3d3} 
jointly leverages jointly multi-camera constraints and monocular depth cues. Most recently, GO-SLAM~\cite{Zhang_2023_ICCV} proposed real-time global pose optimization by considering the complete history of input frames and continuously aligning all poses that enables instantaneous loop closures and correction of global structure.
Still, all SLAM methods assume that the input consists of a sequence of closely related images, \eg with identical intrinsics, nearby camera poses and small illumination variations.
In comparison, our approach handles completely unconstrained image collections.

\myParagraph{3D reconstruction from implicit models} has undergone significant advancements, largely fueled by the integration of neural networks~\cite{nerf, nerfingmvs, deepsdf, monosdf, dist}. Earlier approaches~\cite{niemeyer2020differentiable, dist, deepsdf} utilize Multi-Layer Perceptron (MLP) to generate continuous surface outputs with only posed RGB images. Innovations like Nerf~\cite{nerf} and its follow-ups ~\cite{nerf++, gnerf, wang2021neus, co3d, NeO360, Nerf-in-wild} have pioneered density-based volume rendering to represent scenes as continuous 5D functions for both occupancy and color,  showing exceptional ability in synthesizing novel views of complex scenes. To handle large-scale scenes, recent approaches ~\cite{guo2022neural, monosdf, zhu2022nice, zhu2023nicer} introduce geometry priors to the implicit model, leading to much more detailed reconstructions. %
In contrast to the implicit 3D reconstruction, our work focuses on the explicit 3D reconstruction and showcases that the proposed \duster{} can not only have detailed 3D reconstruction but also provide rich geometry for multiple downstream 3D tasks.

\myParagraph{RGB-pairs-to-3D} takes its roots in two-view geometry~\cite{2003multiple} and is considered as a stand-alone task or an intermediate step towards the multi-view reconstruction. This process typically involves estimating a dense depth map and determining the relative camera pose from two different views. Recent learning-based approaches formulate this problem either as pose and monocular depth regression~\cite{zhou2017unsupervised, yin2018geonet, ranjan2019competitive} or pose and stereo matching~\cite{demon, tang2018ba, deepv2d, wang2021deep, zhao23geofill}. The ultimate goal is to achieve 3D reconstruction from the predicted geometry~\cite{agarwala22plainformers}. In addition to reconstruction tasks, learning from two views also gives an advance in unsupervised pretraining; the recently proposed CroCo~\cite{croco,croco_v2} introduces a pretext task of cross-view completion from a large set of image pair to learn 3D geometry from unlabeled data and to apply this learned implicit representation to various downstream 3D vision tasks. Our method draws inspiration from the CroCo pipeline, but diverges in its application. Instead of focusing on model pretraining, our approach leverages this pipeline to directly generate 3D pointmaps from the image pair. In this context, the depth map and camera poses are only by-products in our pipeline.

\section{Multi-view Pose Estimation}
\label{sec:mvpose}
We include additional results for the multi-view pose estimation task from the main paper (in Sec.~4.2). 
Namely, we compute the pose accuracy for a smaller number of input images (they are randomly selected from the entire test sequences).
\cref{tab:multiview2} reports our performance and compares with the state of the art.
Numbers for state-of-the-art methods are borrowed from the recent PoseDiffusion~\cite{posediffusion} paper's tables and plots, hence some numbers are only approximate.
Our method consistently outperforms all other methods on the CO3Dv2 dataset by a large margin, even for small number of frames.
As can be observed in \cref{fig:opposite} and in the \supplvideo{}, \duster{} handles opposite viewpoints (\ie nearly 180$^\circ$ apart) seemingly without much troubles.
In the end, \duster{} obtains relatively stable performance, regardless of the number of input views.
When comparing with PoseDiffusion~\cite{posediffusion} on RealEstate10K, we report performances with and without training on the same dataset.
Note that \duster{}'s training data include a small subset of CO3Dv2 (we used 50 sequences for each category, \ie less than 7\% of the full training set) but \emph{no data} from RealEstate10K whatsoever.

An example of reconstruction on RealEstate10K is shown in~\cref{fig:realestate}.
Our network outputs a consistent pointcloud despite wide baseline viewpoint changes between the first and last pairs of frames.

\begin{table}[h] \centering
\small
\resizebox{1\linewidth}{!}{
\setlength{\tabcolsep}{2pt}
\begin{tabular}{lcccccc}
\hline
\specialrule{1.5pt}{0.5pt}{0.5pt}
\multirow{2}{*}{Methods} & \multirow{2}{*}{N Frames} & \multicolumn{3}{c}{Co3Dv2~\cite{co3d}} &  & RealEstate10K~\cite{realestate10K}\\ \cline{3-5} \cline{7-7}
                       &  & RRA@15  & RTA@15 & mAA(30) &  & mAA(30)       \\ \specialrule{1.5pt}{0.5pt}{0.5pt}
COLMAP+SPSG     & 3 &  $\sim$22  &  $\sim$14    &  $\sim$15    &  &    $\sim$23        \\
PixSfM     & 3 &  $\sim$18  &  $\sim$8    &  $\sim$10    &  &    $\sim$17        \\
Relpose     & 3 &  $\sim$56  &  -    &  -    &  &    -        \\
PoseDiffusion     & 3 &  $\sim$75  &  $\sim$75    &  $\sim$61    &  &   -\ ($\sim$77)        \\
\bf{\duster \ 512}     & 3 & 95.3    & \bf{88.3}   & \bf{77.5}    &  & \bf{69.5}         \\
\specialrule{1.5pt}{0.5pt}{0.5pt}
COLMAP+SPSG     & 5 &  $\sim$21  &  $\sim$17    &  $\sim$17    &  &    $\sim$34        \\
PixSfM     & 5 &  $\sim$21  &  $\sim$16    &  $\sim$15    &  &    $\sim$30        \\
Relpose     & 5 &  $\sim$56  &  -    &  -    &  &    -        \\
PoseDiffusion     & 5 &  $\sim$77  &  $\sim$76    &  $\sim$63    &  &    -\ ($\sim$78)        \\
\bf{\duster \ 512}     & 5 & 95.5    & 86.7   & 76.5    &  & 67.4          \\
\specialrule{1.5pt}{0.5pt}{0.5pt}
COLMAP+SPSG     & 10 &  31.6  &  27.3    &  25.3    &  &    45.2        \\
PixSfM     & 10 &  33.7  &  32.9    &  30.1    &  &   49.4        \\
Relpose     & 10 &  57.1  &  -    &  -    &  &    -        \\
PoseDiffusion     & 10 &  80.5  &  79.8    &  66.5    &  &    48.0 ($\sim$80)        \\
\bf{\duster \ 512}     & 10 & \bf{96.2}    & 86.8   & 76.7    &  & 67.7          \\ \specialrule{1.5pt}{0.5pt}{0.5pt}
\end{tabular}
}
\caption{Comparison with the state of the art for multi-view pose regression on the CO3Dv2~\cite{co3d} and RealEstate10K~\cite{realestate10K} with 3, 5 and 10 random frames. 
(Parentheses) indicates results obtained after training on RealEstate10K.
In contrast, we report results for \duster{} after global alignment \emph{without} training on RealEstate10K.
\label{tab:multiview2}}
\normalsize
\end{table}

\section{Visual localization}
\label{sec:visloc}

We include additional results of visual localization on the 7-scenes and Cambridge-Landmarks datasets~\cite{7scenes,cambridge}.
Namely, we experiment with a scenario where the focal parameter of the querying camera is unknown.
In this case, we feed the query image and a database image into \duster{}, and get an un-scaled 3D reconstruction.
We then scale the resulting pointmap according to the ground-truth pointmap of the database image, and extract the pose as described in~\cref{ssec:postproc} of the main paper.
\cref{tab:vl_suppl} shows that this method performs reasonably well on the 7-scenes dataset, where the median translation error is on the order of a few centimeters.
On the Cambridge-Landmarks dataset, however, we obtain considerably larger errors. 
After inspection, we find that the ground-truth database pointmaps are sparse, which prevents any reliable scaling of our reconstruction.
On the contrary, 7-scenes provides dense ground-truth pointmaps.
We conclude that further work is necessary for "in-the-wild" visual-localization with unknown intrinsics.

\begin{table*}[t!] 
\begin{center}
\small
\renewcommand\arraystretch{1.2}
\resizebox{1.0\textwidth}{!}{
\setlength{\tabcolsep}{2pt}

\begin{tabular}{llcccccccccccccc}
\specialrule{1.5pt}{0.5pt}{0.5pt} 
\multicolumn{2}{l}{\multirow{2}{*}{Methods}} & GT & \multicolumn{7}{c}{7Scenes (Indoor)~\cite{7scenes}}                                              &  & \multicolumn{5}{c}{Cambridge (Outdoor)~\cite{cambridge}}                       \\ \cline{4-10} \cline{12-16} 
\multicolumn{2}{l}{} & Focals & Chess     & Fire      & Heads     & Office    & Pumpkin   & Kitchen   & Stairs    &  & S. Facade & O. Hospital & K. College & St.Mary’s & G. Court   \\ 
\specialrule{1pt}{0pt}{0pt} 
&{\bf \duster \ 512 from 2D-matching}              & $\checkmark$ & 3/0.97 & 3/0.95 & 2/1.37 & \textbf{3}/1.01 & \textbf{4}/1.14 & 4/1.34 & 11/2.84 &  & 6/0.26 & 17/0.33   & \textbf{11/0.20}& \textbf{7/0.24} & 38/0.16  \\
\hline
&{\bf \duster \ 512 from scaled rel-pose}              & $\times$ & 5/1.08 & 5/1.18 & 4/1.33 & 6/1.05 & 7/1.25 & 6/1.37 & 26/3.56 &  & 64/0.97 & 151/0.88   & 102/0.88 & 79/1.46 & 245/1.08  \\ \specialrule{1.5pt}{0.5pt}{0.5pt} 
\end{tabular}
}

\caption{Absolute camera pose on 7Scenes~\cite{7scenes} (top 1 image) and Cambridge-Landmarks~\cite{cambridge} (top 20 images) datasets. We report the median translation and rotation errors ($cm/ ^{\circ}$).}
\label{tab:vl_suppl}
\normalsize
\end{center}
\end{table*}

\section{Training details}
\label{sec:training_details}
\subsection{Training data}

\myParagraph{Ground-truth pointmaps.}
Ground-truth pointmaps $\Xgt{1}{1}$ and $\Xgt{2}{1}$ for images $I^1$ and $I^2$, respectively, from Eq.~(2) in the main paper are obtained from the ground-truth camera intrinsics $K_1, K_2\in\mathbb{R}^{3\times 3}$, camera poses $P_1, P_2\in\mathbb{R}^{3\times 4}$ and depthmaps $D_1,D_2\in\mathbb{R}^{W\times H}$.
Specifically, we simply project both pointmaps in the reference frame of $P_1$:
\begin{align}
 \Xgt{1}{1} & = K_1^{-1} ([U;V;\mathbf{1}]\cdot D_1) \\
 \Xgt{2}{1} & = P_1 P_2^{-1} h\left( \Xgt{2}{2} \right) \nonumber \\
            & = P_1 P_2^{-1} h\left( K_2^{-1} ([U;V;\mathbf{1}]\cdot D_2) \right),
\end{align}
where $X\cdot Y$ denotes element-wise multiplication, $U,V \in \mathbb{R}^{W \times H}$ are the $x,y$ pixel coordinate grids and $h$ is the mapping to homogeneous coordinates, see Eq.~(1) of the main paper.

\myParagraph{Relation between depthmaps and pointmaps.}
As a result, the depth value $D^1_{i,j}$ at pixel $(i,j)$ in image $I^1$ can be recovered as
\begin{equation}
  D^1_{i,j} = \Xgt{1}{1}_{i,j,2}.
  \label{eq:depthmap}
\end{equation}
Therefore, all depthmaps displayed in the main paper and this \supplementary{} are straightforwardly extracted from \duster{'s} output as $\X{1}{1}_{:,:,2}$ and $\X{2}{2}_{:,:,2}$ for images $I^1$ and $I^2$, respectively.

\myParagraph{Dataset mixture.}
\duster{} is trained with a mixture of eight datasets:
Habitat~\cite{Savva_2019_ICCV}, ARKitScenes~\cite{arkitscenes}, MegaDepth~\cite{megadepth}, Static Scenes 3D~\cite{robust_mvd}, Blended MVS~\cite{blendedMVS}, ScanNet++~\cite{scannet++}, CO3Dv2~\cite{co3d} and Waymo~\cite{Sun_2020_CVPR}. These datasets feature diverse scene types: indoor, outdoor, synthetic, real-world, object-centric, etc. 
Table~\ref{tab:dataset_mix} shows the number of extracted pairs in each datasets, which amounts to 8.5M in total.

\label{ssec:training}
\myParagraph{Data augmentation.} 
We use standard data augmentation techniques, namely random color jittering and random center crops, the latter being a form of focal augmentation. 
Indeed, some datasets are captured using a single or a small number of camera devices, hence many images have practically the same intrinsic parameters.
Centered random cropping thus helps in generating more focals.
Crops are centered so that the principal point is always centered in the training pairs.
At test time, we observe little impact on the results when the principal point is not exactly centered. %
During training, we also systematically feed each training pair $(I^1,I^2)$ 
as well as its inversion $(I^2,I^1)$ to help generalization.
Naturally, tokens from these two pairs do not interact.

\subsection{Training hyperparameters}
\label{sup:implem_details}

We report the detailed hyperparameter settings we use for training \duster{} in Table~\ref{tab:training_step234}.

\begin{table*}[]
    \centering
    \resizebox{\linewidth}{!}{
    \begin{tabular}{l@{\hskip 1.0cm}l@{\hskip 0.8cm}l@{\hskip 0.8cm}l}
    \specialrule{1.5pt}{0.5pt}{0.5pt} 
    Hyperparameters & low-resolution training & high-resolution training & DPT training \\
    \midrule
    Prediction Head & Linear & Linear & DPT\cite{dpt} \\
    \midrule
    Optimizer & AdamW~\cite{adamw} & AdamW~\cite{adamw} & AdamW~\cite{adamw} \\
    Base learning rate & 1e-4 & 1e-4 & 1e-4 \\
    Weight decay & 0.05 & 0.05 & 0.05 \\
    Adam $\beta$ & (0.9, 0.95) & (0.9, 0.95) & (0.9, 0.95) \\
    Pairs per Epoch & 700k & 70k & 70k \\
    Batch size & 128 & 64 & 64 \\
    Epochs & 50 & 100 & 90 \\
    Warmup epochs & 10 & 20 & 15 \\
    Learning rate scheduler & Cosine decay & Cosine decay & Cosine decay \\
    \specialrule{1pt}{0pt}{0pt} 
    \multirow{3}{*}{Input resolutions} & $224{\times}224$ & $512{\times}384$, $512{\times}336$  & $512{\times}384$, $512{\times}336$ \\
                                      & & $512{\times}288$, $512{\times}256$ & $512{\times}288$, $512{\times}256$ \\
                                      & & $512{\times}160$ & $512{\times}160$ \\
    \cmidrule{2-4}
    Image Augmentations & Random centered crop, color jitter & Random centered crop, color jitter  & Random centered crop, color jitter \\
    \cmidrule{2-4}
    Initialization & CroCo v2\cite{croco_v2} & low-resolution training & high-resolution training \\
    \specialrule{1.5pt}{0.5pt}{0.5pt} 
    \end{tabular}
    }
    \vspace{-0.3cm}
    \caption{\textbf{Detailed hyper-parameters} for the training, with first a low-resolution training with a linear head followed by a higher-resolution training still with a linear head and a final step of higher-resolution training with a DPT head, in order to save training time}
    \label{tab:training_step234}
\end{table*}

\begin{table}[h] \centering
\small
\setlength{\tabcolsep}{2pt}
\begin{tabular}{lcr}
\hline
\specialrule{1.5pt}{0pt}{0pt} 
Datasets                      & Type            & N Pairs \\ 
\specialrule{1.5pt}{0.5pt}{0.5pt} 
Habitat~\cite{Savva_2019_ICCV}& Indoor / Synthetic&  1000k\\
CO3Dv2~\cite{co3d}            & Object-centric  &   941k \\
ScanNet++~\cite{scannet++}    & Indoor / Real        &   224k \\
ArkitScenes~\cite{arkitscenes}& Indoor / Real          &  2040k \\
Static Thing 3D~\cite{robust_mvd}& Object / Synthetic&   337k\\ 
MegaDepth~\cite{megadepth}    & Outdoor / Real        &  1761k \\
BlendedMVS~\cite{blendedMVS}  & Outdoor / Synthetic       &  1062k \\
Waymo~\cite{Sun_2020_CVPR}    & Outdoor / Real         &  1100k \\ 
\specialrule{1.5pt}{0.5pt}{0.5pt} 
\end{tabular}
\caption{Dataset mixture and sample sizes for {\duster } training.}
\label{tab:dataset_mix}
\normalsize
\end{table}

{\small
\bibliographystyle{ieee_fullname}
\bibliography{macros,bib}
}

\end{document}